\documentclass{article}
    \usepackage[final]{neurips_2025}

\usepackage{wrapfig}
\usepackage{graphicx}
\usepackage{float}
\usepackage{amsmath}
\usepackage{multirow}
\usepackage{bbm}
\usepackage{listings}
\usepackage{pifont}
\usepackage[utf8]{inputenc} 
\usepackage[T1]{fontenc}    
\usepackage{booktabs}       
\usepackage{amsfonts}       
\usepackage{nicefrac}       
\usepackage{microtype}      
\usepackage[ruled,vlined]{algorithm2e} 
\usepackage[table]{xcolor}
\usepackage{amssymb}       
\usepackage{titletoc}

\usepackage{url}
\usepackage{colortbl}
\usepackage{array} 
\usepackage{mathtools}
\usepackage{bm}
\usepackage{xspace}
\usepackage{titlesec} 

\definecolor{rose}{HTML}{EA018C}
\definecolor{cvprblue}{rgb}{0.21,0.49,0.74}
\definecolor{oxfordblue}{RGB}{0,33,71}
\definecolor{oxfordroyalblue}{RGB}{29,66,166}
\usepackage[colorlinks=true,
    citecolor=cvprblue,
    filecolor=cvprblue,
    urlcolor=rose]{hyperref}

\definecolor{my_blue}{HTML}{d3eaf2}
\definecolor{my_red}{HTML}{F8CECC}
\definecolor{my_green}{HTML}{D5E8D4}

\newcommand{\seal}{\raisebox{-0.2\height}{\includegraphics[height=1.5em]{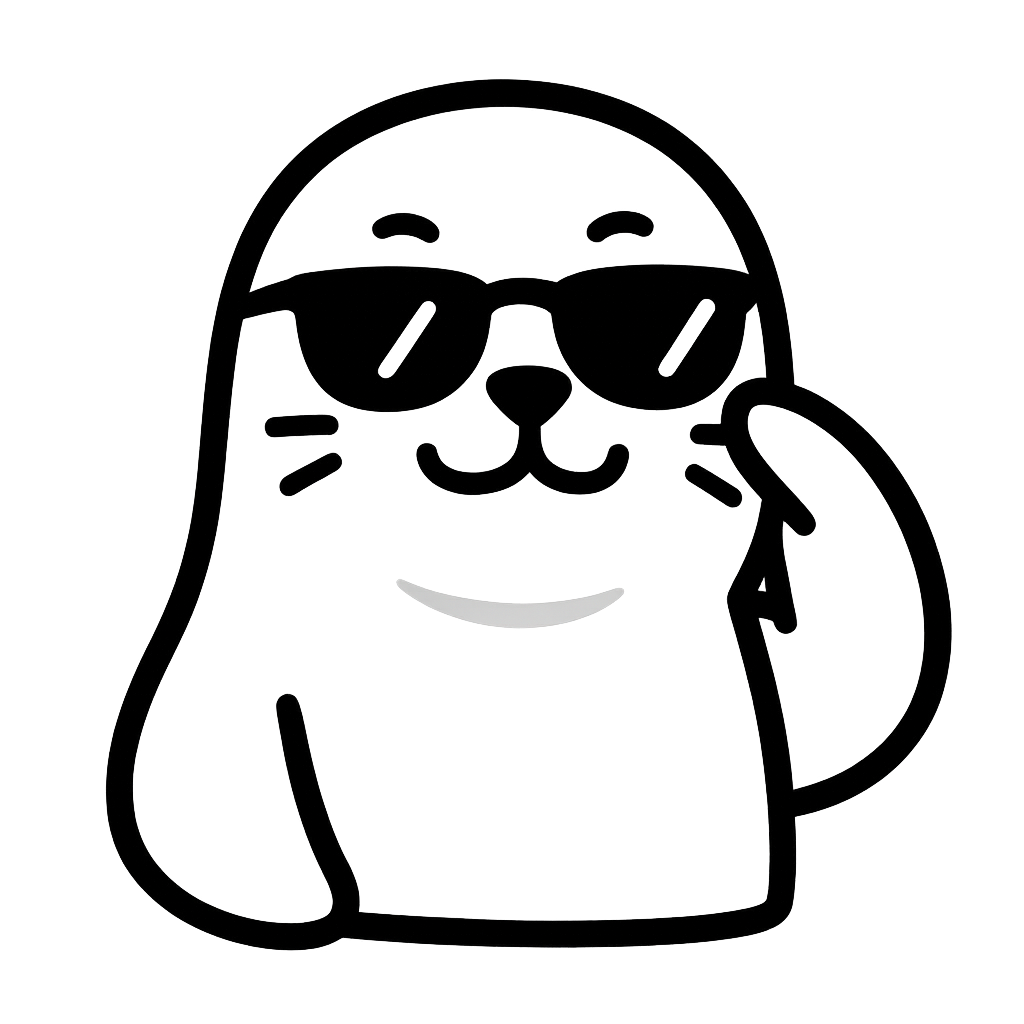}}}

\makeatletter
\def\onedot{\futurelet\@let@token\@onedot}
\def\@onedot{\ifx\@let@token.\else.\null\fi\xspace}
\makeatother
\usepackage{xspace}
\newcommand{\eg}{\emph{e.g.}\xspace}

\title{\seal  SEAL: Semantic-Aware Hierarchical Learning for Generalized Category Discovery}

\author{Zhenqi He\textsuperscript{*} \qquad Yuanpei Liu\textsuperscript{*} \qquad Kai Han\textsuperscript{\textdagger}\\
Visual AI Lab, The University of Hong Kong \\
{\tt\small \{zhenqi\_he, ypliu0\}@connect.hku.hk \qquad  kaihanx@hku.hk}}

\begin{document}
\renewcommand{\thefootnote}{\fnsymbol{footnote}}
\footnotetext[1]{Equal contribution.}
\footnotetext[2]{Corresponding author.}
\maketitle

\begin{abstract}
This paper investigates the problem of Generalized Category Discovery (GCD). 
Given a partially labelled dataset, GCD aims to categorize all unlabelled images, regardless of whether they belong to known or unknown classes.
Existing approaches typically depend on either single-level semantics or manually designed abstract hierarchies, which limit their generalizability and scalability.
To address these limitations, we introduce a \textbf{SE}mantic-aware hier\textbf{A}rchical \textbf{L}earning framework (\textbf{SEAL}), guided by naturally occurring and easily accessible hierarchical structures.
Within SEAL, we propose a Hierarchical Semantic-Guided Soft Contrastive Learning approach that exploits hierarchical similarity to generate informative soft negatives, addressing the limitations of conventional contrastive losses that treat all negatives equally.
Furthermore, a Cross-Granularity Consistency (CGC) module is designed to align the predictions from different levels of granularity.
SEAL consistently achieves state-of-the-art performance on fine-grained benchmarks, including the SSB benchmark, Oxford-Pet, and the Herbarium19 dataset, and further demonstrates generalization on coarse-grained datasets.
Project page: \url{https://visual-ai.github.io/seal/}
\end{abstract}
\section{Introduction}
The field of computer vision has undergone substantial progress in various tasks, including classification~\cite{simonyan2014very, he2016deep}, object detection~\cite{girshick2015fast, ren2015faster}, and segmentation~\cite{he2017mask, transnuseg, wang2020solo}.
Such advancements have largely been driven by access to large-scale, human-annotated datasets~\cite{deng2009imagenet, lin2014microsoft}.
However, models trained on these datasets are constrained to a closed-world paradigm, limiting their predictions to the predefined labels within the training set.
In contrast, there exists a wealth of unlabelled data in the open world. To capitalize on the unlabelled data, a variety of Semi-Supervised Learning (SSL) techniques~\cite{chapelle2009semi}  have been proposed, yielding notable improvements over traditional supervised learning methods. 
Despite substantial success in various tasks~\cite{berthelot2019mixmatch,xu2021end,chen2021semi}, most existing SSL methods are designed under the closed-set assumption, wherein the training and test datasets share an identical set of classes. 
Category discovery, initially introduced as Novel Category Discovery (NCD)~\cite{han2019learning,he2025category} and later extended to Generalized Category Discovery (GCD)~\cite{vaze2022generalized,he2025category}, has recently emerged as a compelling open-world problem, attracting significant attention. 
Unlike SSL, GCD tackles the challenges where the unlabelled subset may include instances from both known and unknown classes. 
Its primary objective is to utilise knowledge gained from labelled data to effectively categorize all samples within the unlabelled data. 
Concurrently, an equivalent task named Open-world Semi-Supervised Learning (OSSL)~\cite{cao2022open} has also been introduced.

\begin{figure*}[t]
\centering
\includegraphics[width = \textwidth]{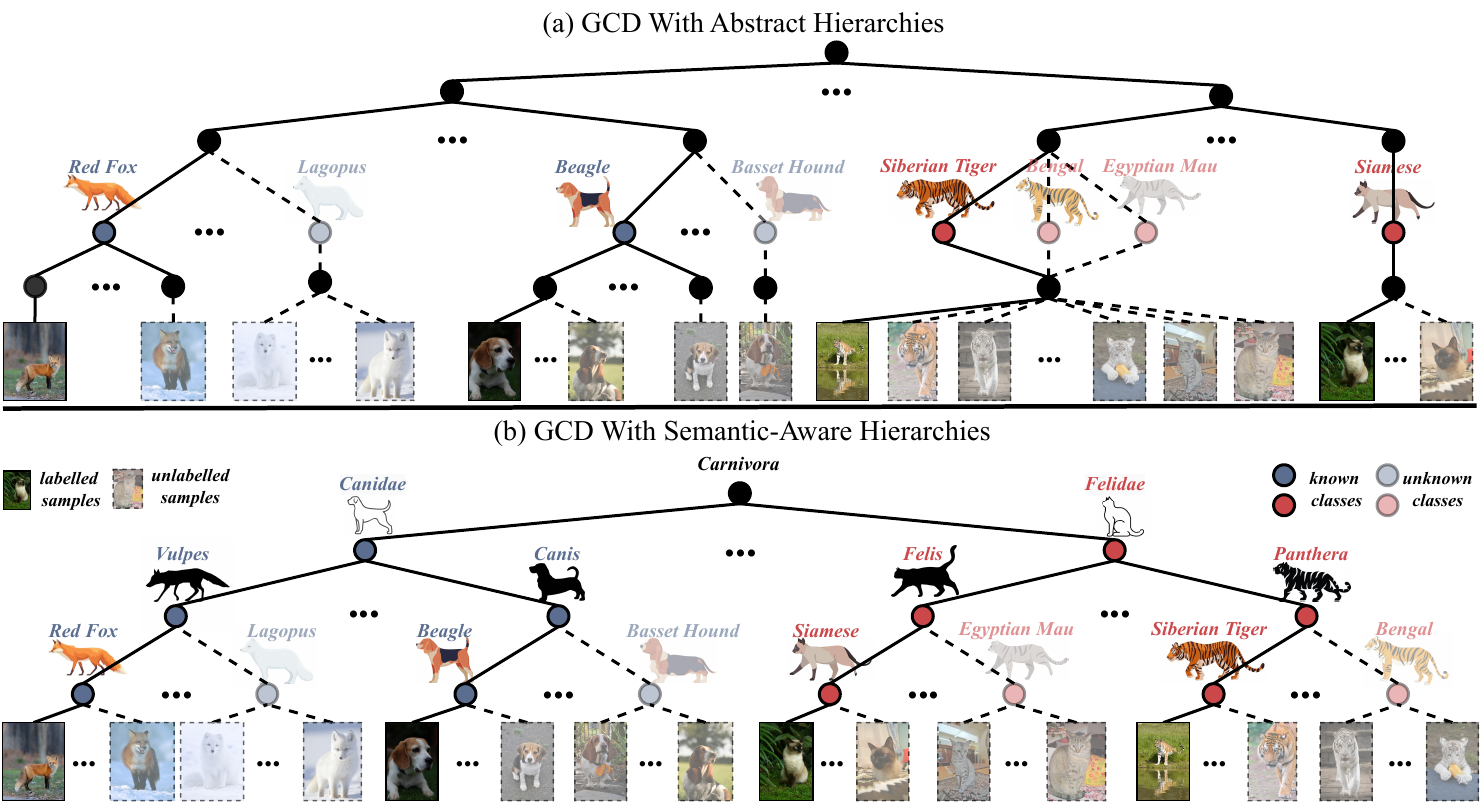}
\caption{
Comparison of SEAL and previous methods~\cite{rastegar2023learn,wang2023discover} using hierarchical learning. (a) In previous attempts, several upper and lower levels as well as abstract concepts are defined around the ground-truth level, which may cause errors in the hierarchical structure. (b) In our method, we propose to utilise the semantic information at different levels to enhance the GCD performance.}
\label{fig:intro}
\end{figure*}

The effectiveness of GCD is rooted in the efficient \textit{transfer} of knowledge from known categories to \textit{cluster} samples of both known and novel categories.
As a \textit{transfer clustering} task~\cite{han2019learning}, hierarchical information has been demonstrated to be effective in GCD~\cite{rastegar2023learn, hao2023cipr} and similarly in the parallel task of OSSL~\cite{wang2023discover}, particularly with fine-grained datasets~\cite{vaze2022semantic}.
In~\cite{rastegar2023learn}, the hierarchical structure is composed of \textit{abstract concepts} as an implicit binary tree, where each node represents an increasingly abstract concept derived from shared binary code prefixes, and in~\cite{RastegarECCV2024}, the hierarchical structure is implicitly formed by incrementally halving the category count as the hierarchy level increases with hierarchical pseudo-labeling to provide soft supervision for the training.
Similarly, CiPR~\cite{hao2023cipr} constructs abstract hierarchies by iteratively merging data partitions through semi-supervised clustering.
%
The hierarchical tree in~\cite{wang2023discover} consists of manually defined upper and lower levels that represent different granularities, with the number of categories per level controlled by hyperparameters.
More recently, HypCD~\cite{liu2025hyperbolic} implicitly models hierarchies via hyperbolic embeddings, achieving strong performance and underscoring the importance of hierarchical information for GCD.
%
%
These methods build hierarchical levels from abstract, weakly supervised structures that may introduce noise and errors, ultimately affecting GCD performance.
As illustrated in Fig.~\ref{fig:intro} (a), the \textit{`Siberian Tiger'}, \textit{`Bengal'} and \textit{`Egyptian Mau'} can be merged into a single category while the \textit{`Red Fox'} can be divided into multiple categories. 
Additionally, the high similarity among categories may result in some images of the \textit{`Basset Hound'} being incorrectly merged with those of the \textit{`Beagle'}.
This observation naturally prompts us to consider: \textit{whether the intrinsic, semantically grounded taxonomies present in the real world can serve as more reliable guides}.
In botanical research, taxonomists commonly use labelled specimens of known species to classify newly collected, unlabelled samples into existing taxonomic hierarchies or to identify unseen species~\cite{mace2004role, karbstein2024species}.
Similarly, studies in closed-world visual classification~\cite{Chang2021Labrador, 9609630, du2021progressive} have shown that hierarchical structures enhance classification. From an information-theoretic perspective, we further deduce that such semantic-aware hierarchies yield a tighter mutual information bound, providing a principled foundation for our design.

%

To this end, we propose the \textbf{SE}mantic-aware hier\textbf{A}rchical \textbf{L}earning (\textbf{SEAL}) framework for GCD. Unlike previous approaches that either focus exclusively on single-granularity information~\cite{vaze2022generalized, vaze2023no, wang2024sptnet, wen2023parametric, liu2025debgcd} or rely on abstract hierarchical cues~\cite{rastegar2023learn, hao2023cipr, wang2023discover, RastegarECCV2024, liu2025hyperbolic}, SEAL effectively leverages the naturally occurring semantic hierarchies without manual design (shown in Fig.~\ref{fig:intro} (b)) and incorporates several innovative techniques tailored specifically for this task.
\textit{Firstly}, we implement a multi-task training paradigm that facilitates the simultaneous discovery of categories across several different semantic levels.
\textit{Secondly}, we introduce a Cross-Granularity Consistency (CGC) module to align the class predictions from different granularities.
\textit{Thirdly}, we propose the Hierarchical Semantic-guided Soft Contrastive Learning to capture uncertainty in contrastive learning, ensuring that not all negative samples are treated equally.
By effectively integrating these components into a cohesive framework, SEAL can be trained end-to-end in a single stage. 
%


In summary, we make the following contributions in this paper:
\textbf{(i)} We propose \textbf{SEAL}, a novel framework specifically designed to tackle the challenging GCD task by leveraging the inherent semantic hierarchies, marking the first exploration of this aspect. 
\textbf{(ii)} Within the SEAL framework, we develop two novel components: the Cross-Granularity Consistency (CGC) module and Hierarchical Semantic-guided Soft Contrastive Learning. These components function synergistically to significantly enhance the model's category discovery capabilities.
\textbf{(iii)} Through extensive experimentation on public GCD benchmarks, SEAL consistently demonstrates its effectiveness and achieves superior performance, especially on fine-grained datasets. 
\section{Related Work}
\paragraph{Category Discovery.}
Novel Category Discovery (NCD) was first articulated in~\cite{han2019learning}, establishing a pragmatic framework for transferring knowledge from known categories to clusters of unseen categories, framed as a transfer clustering problem.
Subsequently, a variety of methods have emerged to advance the research domain~\cite{han2020automatically, han2021autonovel, jia2021joint, zhao2021novel, ncl, fini2021unified}.
Generalized Category Discovery (GCD) extends the NCD framework by relaxing its assumptions, incorporating unlabelled data that features samples from both known and unknown classes~\cite{vaze2022generalized}.
Recent studies~\cite{cao2022open, hao2023cipr, pu2023dynamic, joseph2022novel, cendra2024promptccd, wang2024hilo, liu2025debgcd} have explored a range of strategies to tackle the challenges introduced by GCD.
Notably, InfoSieve~\cite{rastegar2023learn} and CiPR~\cite{hao2023cipr} guide category discovery using abstract hierarchies that are automatically inferred from the data. A similar approach is employed in the OSSL task by TIDA~\cite{wang2023discover}, which employs handcrafted abstract hierarchies by constructing prototypes at manually defined levels.
Conversely, SimGCD~\cite{wen2023parametric} introduces an entropy-regularized classifier that provides a robust baseline. SPTNet~\cite{wang2024sptnet} builds upon SimGCD by incorporating spatial prompt tuning to emphasize salient object parts, while DebGCD~\cite{liu2025debgcd} proposes a distribution-guided debiased learning framework to address the inherent label bias and semantic shifts in GCD.

\paragraph{Hierarchical Learning.}
In the realm of hierarchical learning, numerous studies~\cite{Chang2021Labrador, 7583684, 9609630, du2021progressive} have explored the use of hierarchical label information to enhance classification performance, particularly in closed-world settings. For instance, \cite{Chang2021Labrador} employs a multi-task framework that utilises coarse-to-fine labels to improve fine-grained recognition, whereas \cite{hierarchicalContrastiveLearning} introduces hierarchical contrastive learning to enrich representations with multi-level semantic cues. More recently, hierarchical learning has been adapted to open-set recognition, as demonstrated in~\cite{lang2024osr, Wu_2024_CVPR}, where semantic hierarchies contribute to improved generalization to unseen classes. To the best of our knowledge, our work is the first to apply semantic-guided hierarchies to the GCD task, facilitating the effective discovery and classification of both known and novel categories.
\section{Preliminary}
\label{Sec:pre}
\textbf{Problem Statement:} GCD aims to develop models capable of classifying unlabelled samples from known categories while simultaneously clustering those from unknown categories. Formally, we are given a labelled dataset $\mathcal{D}_l = {(\bm{x}^{l}_{i}, y^{l}_{i})} \subset \mathcal{X} \times \mathcal{Y}_l$ and an unlabelled dataset $\mathcal{D}_u = {(\bm{x}^{u}_{i}, y^{u}_{i})} \subset \mathcal{X} \times \mathcal{Y}_u$, where $\mathcal{Y}_l \subset \mathcal{Y}_u$. The unlabelled data includes samples from both known and novel categories. The number of known categories is denoted by $M = |\mathcal{Y}_l|$, and the total number of categories is $K = |\mathcal{Y}_l \cup \mathcal{Y}_u|$. Following prior works~\cite{han2021autonovel,wen2023parametric,vaze2023no}, we assume $K$ is known during training. When $K$ is unknown, it can be estimated using techniques such as~\cite{han2019learning,vaze2022generalized}.

\textbf{Revisiting Baseline:}
SimGCD~\cite{wen2023parametric} is a representative end-to-end baseline for GCD that unifies contrastive representation learning and parametric classification. The model employs a Vision Transformer~\cite{dosovitskiy2020image} backbone pretrained using DINO~\cite{caron2021emerging}, where the input image $\boldsymbol{x}_i$ is first passed through an embedding layer $\varphi$ and the feature extractor $\mathcal{F}$, followed by a projection head $\mathcal{H}$ to produce a normalized representation $\mathbf{z}_i = \mathcal{H}(\mathcal{F}(\varphi(\boldsymbol{x}_i)))/ |\mathcal{H}(\mathcal{F}(\varphi(\boldsymbol{x}_i)))|$. The representation learning objective $\mathcal{L}_{rep}$ is based on the InfoNCE loss~\cite{oord2018representation}:
\begin{equation}
\mathcal{L}_{rep}(\boldsymbol{x}_i) = -\frac{1}{|\mathcal{P}(\boldsymbol{x}_i)|} \sum_{\mathbf{z}_i^{+} \in \mathcal{P}(\boldsymbol{x}_i)} \log \sigma(\mathbf{z}_i \cdot \mathbf{z}_i^{+}; \tau),
\end{equation}
where $\mathcal{P}(\boldsymbol{x}_i)$ denotes the set of positive features (\eg, different views of the same image), and $\sigma(\cdot;\tau)$ is the softmax with temperature $\tau$. For labelled samples, additional positives from the same class are used to enable supervised contrastive learning.

For the parametric classification, SimGCD adopts a cosine-based classifier~\cite{gidaris2018dynamic} with a learnable prototype set $\mathcal{C}=\{\bm{c}_1,...,\bm{c}_K\}$ where each prototype $\bm{c}_k$ is $l_2$-normalized and the output probability for the $k$-th category is given by ${\bm{p}_i}^{(k)}=\sigma(\mathbf{z}_i \cdot \bm{c}_k; \tau)$. Given the pseudo-label $\boldsymbol{q}_i$ obtained from a sharpened prediction of a different view, the classification loss is:
\begin{equation}
    \mathcal{L}_{cls}^{u}=\frac{1}{|\mathcal{B}|}\sum_{i\in \mathcal{B}} l_{ce}(\bm{q}_i,\bm{p}_i)-\xi H(\overline{\bm{p}}), 
\end{equation}
where $\mathcal{B}$ is current image batch and $H(\overline{\bm{p}})$ denotes the entropy of the mean prediction $\overline{\bm{p}}$. Specifically, for each $\boldsymbol{x}_i$ in the labelled batch $\mathcal{B}_l$, an additional  $\boldsymbol{y}_i$ as the one-hot ground-truth vector is also used for supervised classification loss written as $\mathcal{L}_{cls}^s=\frac{1}{|\mathcal{B}_l|}\sum\nolimits_{i\in \mathcal{B}_l}l_{ce}(\bm{p}_i,\bm{y}_i)$. 
Then, the overall classification loss is formulated as $\mathcal{L}_{cls}=(1-\lambda_b )\mathcal{L}_{cls}^u+\lambda_b \mathcal{L}_{cls}^s$ where $\lambda_b$ is a balance factor.
The final training objective combines both representation and classification terms: $ \mathcal{L}_{bs} = \mathcal{L}_{cls} + \mathcal{L}_{rep}.$

\section{Method}
Before delving into methodological details, we begin with an intuitive hypothesis that underlies our framework: \textit{Leveraging structured semantic hierarchies across multiple levels can facilitate more informative and robust feature learning for GCD setting.}
To support this intuition, we first present a theoretical justification grounded in information theory, demonstrating that the incorporation of hierarchical labels yields a tighter bound on mutual information. This theoretical insight lays the foundation for the design of our approach, which we detail in the subsequent sections.


\subsection{Theoretical Motivation}

From the perspective of information theory, with denoting model parameter as $\theta$, data as $\mathcal{X}$, and label as $\mathcal{Y}$, we write $Z = f_\theta(\mathcal{X})$ as the \textit{deterministic representation} of $\mathcal{X}$ once model $\theta$ is fixed. The optimisation objective is then to maximize the \textit{mutual information} between $Z$ and $Y$~\cite{boudiaf2020unifying}, which can be re-formulated as $\min_{\theta}\;
\Big\{-I_{\theta}\!\bigl(Z_{l};\,\mathcal{Y}_{l}\bigr)
\;+\;
\beta\,
\Bigl[
H_{\theta}\!\bigl(\hat{{Y}_u}\mid \mathcal{X}_{u}\bigr)
   -H_{\theta}\!\bigl(\hat{{Y}_u}\bigr)
\Bigr]\Big\}$ with detailed proof provided in the Appendix,
where $\hat{{Y}_u}$ is the model prediction for unlabelled data, and $\beta$ is the weight factor. Assuming the coarse-grained semantic hierarchical labels $\mathcal{Y}^{(1)}_l,...,\mathcal{Y}^{(H-1)}_l$ are accessible for all labelled samples, the objective naturally extends to:
\begin{equation}
    \min_{\theta}
   \Big\{-\,I_\theta\!\bigl(Z_l;\,\mathcal{Y}_l^{(1)},\dots ,\mathcal{Y}_l^{(H)}\bigr)
   +\beta \Bigl[ H_\theta\!\bigl(\hat{ Y}^{(1:H)}_u\mid \mathcal{X}_u\bigr)
   -\,H_\theta\!\bigl(\hat{ Y}^{(1:H)}_u\bigr)
\Bigr]\Big\}.
\end{equation}
By applying the \textit{chain rule} of mutual information, the supervised part satisfies:
\begin{equation}
I_\theta\!\bigl(Z_l;\mathcal Y_l^{(1:H)}\bigr)
  = I_\theta\!\bigl(Z_l;\mathcal Y_l^{(H)}\bigr)
     + \sum_{h=1}^{H-1}
       I_\theta\!\bigl(
           Z_l;\mathcal Y_l^{(h)}
           \,\bigm|\,
           \mathcal Y_l^{(h+1)},\dots,\mathcal Y_l^{(H)}
       \bigr) 
  \geq I_\theta\!\bigl(Z_l;\mathcal Y_l^{(H)}\bigr).
\end{equation}

Analogously, for the unsupervised component, we obtain:
\begin{equation}
\begin{aligned}
H_\theta\!\bigl(\hat{ Y}^{(1:H)}_u\mid \mathcal{X}_u\bigr)
\;-\;
H_\theta\!\bigl(\hat{ Y}^{(1:H)}_u\bigr)
&= -\,I_\theta\!\bigl(\mathcal{X}_u;\hat Y^{(H)}_u\bigr)
    \;-\;\sum_{h=1}^{H-1} I_\theta\!\bigl(\mathcal{X}_u;\hat Y^{(h)} \mid \hat Y^{(h+1:H)}_u\bigr) \\[2pt]
&\le\; -\,I_\theta\!\bigl(\mathcal{X}_u;\hat Y^{(H)}_u\bigr)
=-H_{\theta}\!\bigl(\hat{{Y}_u}\bigr)
   +\
   H_{\theta}\!\bigl(\hat{{Y}_u}\mid \mathcal{X}_{u}\bigr).
\end{aligned}
\end{equation}

Combining the supervised and unsupervised parts, we have:
\begin{equation}
\begin{aligned}
-\,I_\theta\!\bigl(Z_l;\,\mathcal{Y}_l^{(1)},\dots ,\mathcal{Y}_l^{(H)}\bigr)
   +\beta \Bigl[ H_\theta\!\bigl(\hat{ Y}^{(1:H)}_u\mid \mathcal{X}_u\bigr)
   -\,H_\theta\!\bigl(\hat{ Y}^{(1:H)}_u\bigr)
\Bigr] \\
\leq -I_{\theta}\!\bigl(Z_{l};\,\mathcal{Y}_{l}^{(H)}\bigr)
\;+\;
\beta\,
\Bigl[
   -H_{\theta}\!\bigl(\hat{{Y}}^{(H)}_u\bigr)
   +\;
   H_{\theta}\!\bigl(\hat{{Y}}^{(H)}_u\mid \mathcal{X}_{u}\bigr)
\Bigr].
\end{aligned}
\end{equation}
Therefore, from the perspective of information theory, incorporating semantic hierarchical labels provides a strictly tighter upper bound on the mutual information, which motivates us to introduce the semantic-guided hierarchical learning framework for GCD.


\subsection{SEAL: Semantic-Aware Hierarchical Learning for GCD}
\begin{figure}[t]
\centering
\includegraphics[width = \textwidth]{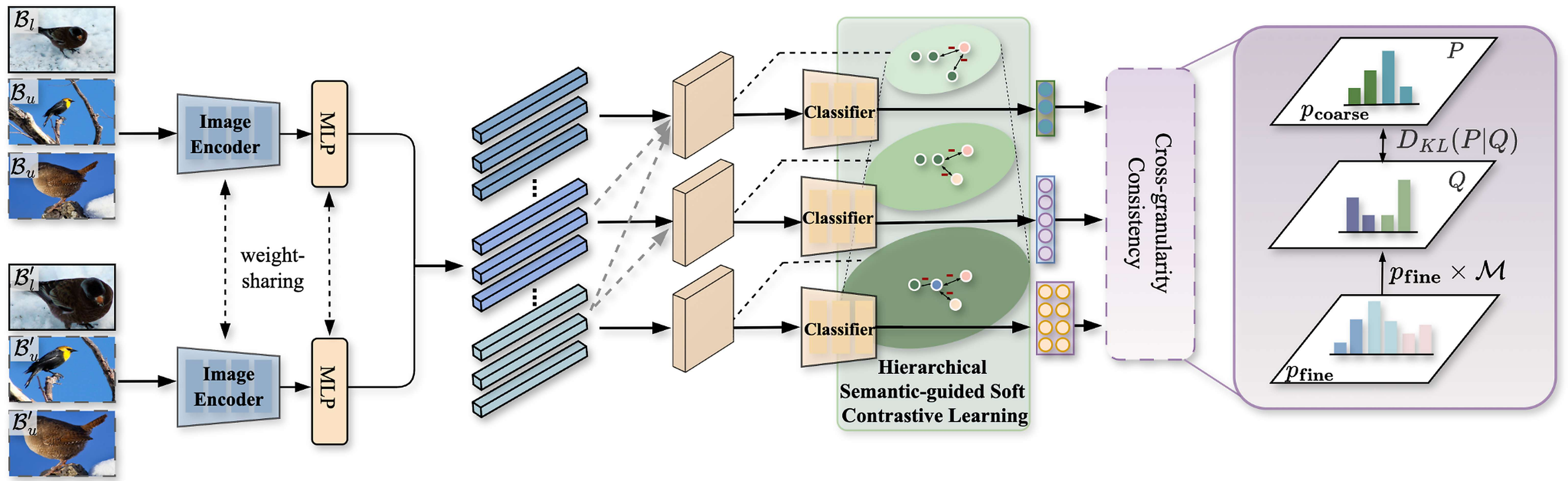}
\caption{
Overview of the proposed SEAL framework. 
}
\label{fig:method}
\end{figure}

Building on the advantages of semantic hierarchies, we propose the \textbf{SE}mantic-aware hier\textbf{A}rchical \textbf{L}earning ({SEAL}) framework for GCD.
The overall framework is outlined in Fig.~\ref{fig:method}. In contrast to prior GCD approaches that either rely solely on single-granularity information~\cite{vaze2022generalized, vaze2023no, wang2024sptnet, wen2023parametric} or depend on abstract hierarchies~\cite{rastegar2023learn, wang2023discover}, we embed explicit semantic structure via three key elements: (1) a semantic-aware multi-task framework; (2) a cross-granularity consistency module to align predictions across levels; and (3) a hierarchical soft contrastive learning strategy to mitigate the ``equivalent negative'' assumption by weighting dissimilarity according to semantic proximity.

\subsubsection{Semantic-aware Hierarchical Learning}
We first introduce the semantic-aware multi-task framework.
Inspired by~\cite{Chang2021Labrador}, we advocate for a joint learning framework across multiple semantic levels, allowing information across the hierarchies to guide and strengthen representation learning at the target granularity.
We define $H$ as the number of semantic levels, with corresponding ground-truth labels  $\bm{y}_1, \ldots, \bm{y}_H$ ordered from coarse to fine.
Our multi-task architecture couples a shared image encoder $\mathcal{F}$ followed by a projection layer $\phi$ to disentangle features for various granularities, which can be formulated as $\mathbf{z} \;=\; \phi\bigl(\mathcal{F}(\mathbf{x})\bigr) 
      \;=\; \bigl[\,\mathbf{z}_1 ; \mathbf{z}_{2} ; \dots ; \mathbf{z}_{H}\,\bigr]$, 
where `$;$' denotes concatenation.
Following the observation in \cite{Chang2021Labrador} that fine-grained features can benefit coarse-grained predictions but not vice versa, we reuse lower-level features when computing coarse-level outputs.
To avoid training bias towards coarse branches, we adopt a gradient controller $\Gamma$ to include fine-level features without allowing gradient backpropagation.
Formally, the aggregated feature of sample $\bm{x}_i$ at the $h$-{th} level is $\hat{\mathbf{z}}_i = [\mathbf{z}_1 ; \cdots ;\mathbf{z}_h;\Gamma(\mathbf{z}_{h+1}); \cdots;\Gamma(\mathbf{z}_H)]$, where $\Gamma(\cdot)$ stops gradient propagation during training. 
We train a GCD classifier at each level. For $h$-{th} level, the classification loss is denoted as $\mathcal{L}_{cls}^h$.

\subsubsection{Cross-Granularity Consistency Self Distillation}
Although multi-level classification has been widely studied in closed-world settings~\cite{Chang2021Labrador, 9609630}, prior methods often treat each semantic level in isolation, leading to inconsistencies such as assigning labels like \textit{`Shiba'} and \textit{`Cat'} at different granularities for the same instance. 
This lack of cross-level interaction weakens the benefits of hierarchical learning. We address this with a Cross-Granularity Consistency (CGC) module that distills information between granularities to keep predictions mutually coherent.
%
%
Concretely, we add a self-distillation term that minimises the KL divergence between the coarse-level posterior $p(\bm{x}_i|\bm{\theta}_h)$ and a pseudo-coarse distribution obtained by mapping the target posterior $p(\bm{x}_i|\bm{\theta}_H)$ where $\bm{\theta}_h$ denotes the model parameters at granularity $h$.
%
Specifically, we define a dynamic transition matrix $M_{h} \in \mathbb{R}^{n_H \times n_h}$ at granularity $h$ where $n_h$ denotes the number of categories at that level. Each row of $M_h$ encodes how a fine-grained class distributes over coarse classes.
For known fine-grained categories, this is a fixed one-hot vector; for novel classes, we initialize with a uniform distribution and iteratively refine it during training (See Algo. 1 Dynamic Update of $M_h$).
%
%
The pseudo-coarse probability thus can be computed as $p(\bm{x}_i|\bm{\theta}_H) \times M_h$ and  the hierarchical consistency loss at level $h$ is defined as $D_{KL}(p(\bm{x}_i|\bm{\theta}_h) | p(\bm{x}_i|\bm{\theta}_H) \times M_h)$. 
Summing across levels, the overall CGC loss becomes:
\begin{equation}
    \mathcal{L}_{cgc} = \sum_{h=1}^{H-1} D_{KL}(p(\bm{x}_i|\bm{\theta}_h) | p(\bm{x}_i|\bm{\theta}_H) \times M_h),
\end{equation}
where $p(\bm{x}_i|\bm{\theta}_h) = \sigma (f_{\bm{\theta}_h}(\bm{x}_i), \tau_c)$ with  $\sigma(\cdot)$ denoting the softmax operation and $\tau_c$  be the consistency temperature and $f_{\bm{\theta}_h}(\bm{x}_i)$ being logits computed for granularity $h$.

\begin{algorithm}[h]
\label{algo:updateM}
\caption{Dynamic Update of $M_h$}
\KwIn{Model $f$, number of class at level $h$, $n_h$, known fine-grained classes $C_{base}$}
\textbf{Dynamic Update:} \\
    Compute logits $l_{h},\, l_{H} = f(\mathcal{D})$ \\
    Compute prediction probability $p_{h},\, p_{H} = \sigma(l_{h}, \tau_c),\, \sigma(l_{H}, \tau_c)$\\
    \For{each fine class index $k$ not in $C_{base}$}{
        ~~Compute fine-grained predictions $\bm{y}_H = \text{argmax}(p_H)$ \\
        Compute average probability distribution for samples predicted as fine class $k$ : \\
        ~~~~~~~~$\text{avg\_h\_prob} = \text{mean}\!\big(p_h[\bm{y}_H == k]\big)$\\
        Momentum update $M_{h}[k]$ as follows: $M_{h}[k] \gets \lambda \cdot M_{h}[k] + (1 - \lambda) \cdot \text{avg\_h\_prob}$\\
        Normalize $M_{h}[k]$: $M_{h}[k] \gets \dfrac{M_{h}[k]}{\sum M_{h}[k]}$
    }
\KwOut{$M_{h}$}
\end{algorithm}

\subsubsection{Hierarchical Semantic-guided Soft Contrastive Learning}
To strengthen the discriminative capacity of representations in GCD, we propose a Hierarchical Semantic-guided Soft Contrastive Learning approach, addressing key limitations of existing contrastive learning approaches.
Prior GCD methods~\cite{han2021autonovel, wen2023parametric, vaze2023no, liu2025debgcd} treat each non-positive in a mini-batch as an equally hard negative, ignoring semantic relatedness. We instead leverage the hierarchy in our multi-level framework to compute similarity-aware targets, assigning softer negative weights to semantically closer samples and preserving full penalties for unrelated ones.
%
%
We compute pairwise similarities within each mini-batch at every semantic level, yielding similarity matrices $S_h$ at the $h$-th granularity, where $S_h = \frac{{\mathbf{Z}}_h \cdot ({\mathbf{Z}}_h)^\top}{\|\mathbf{Z}_h \| \times \|\mathbf{Z}_h^\top \|} \in \mathbb{R}^{B \times B}$ with $\mathbf{Z}_h $ being the features of the mini-batch at granularity $h$ and $B$ being the mini-batch size.
Each fine-level matrix is then fused with its coarser counterpart, yielding a hierarchical similarity matrix $\tilde{S}_h$.
%
We then generate semantic-aware soft labels as a matrix: $\tilde{Y}_{{soft}_h} = (1 - \lambda_s) \cdot \bm{I} + \lambda_s \cdot \tilde{S}_h$, where $\bm{I}$ is the identity matrix and $\lambda_s$ controls the smoothness of the semantic-aware soft labels. The resulting semantic-guided hierarchical soft contrastive loss is defined as:
\begin{equation}
    \mathcal{L}_{hscl}^h = -\frac{1}{|B|} \sum_{i=1}^B \sum_{j=1}^B \tilde{Y}_{soft_h}(i,j)  \log \frac{\exp(sim(\mathbf{z}_i , \mathbf{z}_j^\prime))}{\sum\nolimits_m^{m\neq i}\exp(sim(\mathbf{z}_i, \mathbf{z}_m^\prime))},
\end{equation}
where $sim(\cdot)$ represents the similarity metric between feature $\mathbf{z}_i$ from $\bm{x}_i$ and feature $\mathbf{z}_j^\prime$ from the augmented view of $\bm{x}_j$, and $\tilde{Y}_{soft_h}(i,j)$ refers to the $(i,j)$ element of the soft label matrix.
Unlike prior works~\cite{wen2023parametric, wang2024sptnet} that rely solely on angle or distance-based measure, we adopt a hybrid metric defined as $sim(\mathbf{z}_i,\mathbf{z}_k^\prime ) = \lambda_c  \mathbf{z}_i \cdot
\mathbf{z}_k^{\prime \top} - (1- \lambda_c) \bigl\lVert \frac{\mathbf{z}_i}{\|\mathbf{z}_i\|} - \frac{\mathbf{z}_k^\prime}{\|\mathbf{z}_k^\prime\|} \bigr\rVert_2$ where $\lambda_c$ is the weighting coefficient that linearly gradually decays during training. This design implements a \textit{curriculum learning} strategy: it begins with easier angle-based cues and gradually adds distance terms to refine representations. 
More ablation studies about the decay schedule are in the Appendix.
\begin{figure*}[t]
\centering
\includegraphics[width = \textwidth]{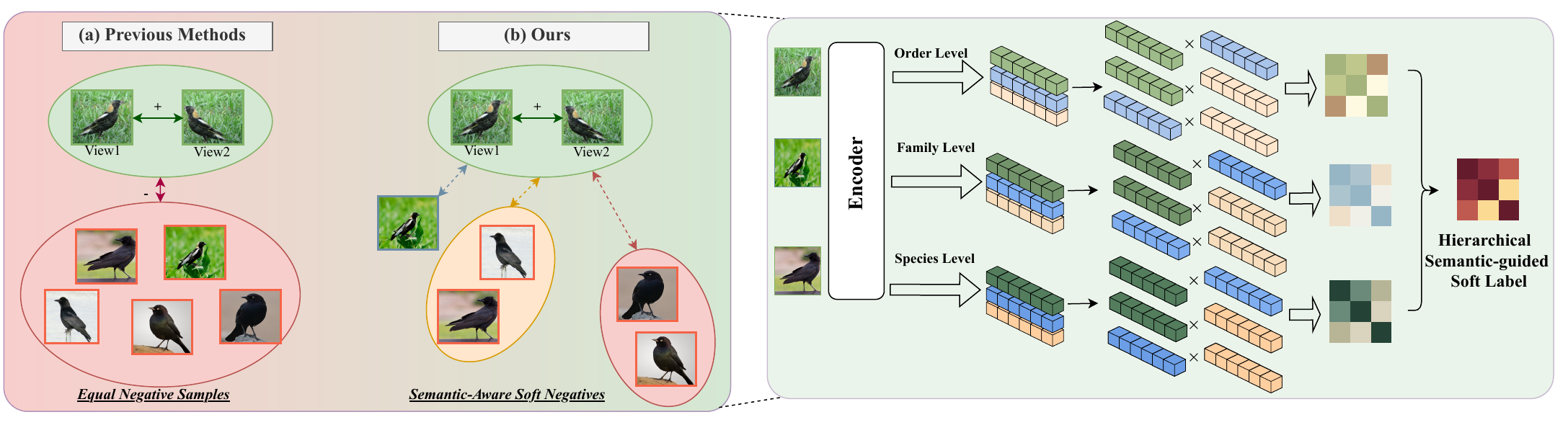}
\caption{
Overview of Hierarchical Semantic-guided Soft contrastive learning. }
\label{fig:soft}
\end{figure*}

\subsubsection{Overall Objective}
Based on the baseline SimGCD~\cite{wen2023parametric} classifier, our framework is designed to be trained in a multi-task manner.
We first replace the original InfoNCE loss~\cite{oord2018representation} in the baseline representation loss $\mathcal{L}_{rep}$ introduced in Sec.~\ref{Sec:pre} by our proposed hierarchical soft contrastive loss $\mathcal{L}_{hscl}^h$, denoting the resulting training objective at each granularity as $\mathcal{L}_{soft_{rep}}^h$.
The final training objective can be formulated as 
\begin{equation}
    \mathcal{L}_{all} = \sum_{h}^H (\mathcal{L}_{soft_{rep}}^h + \mathcal{L}_{cls}^h) + \mathcal{L}_{cgc}
\end{equation}
\section{Experiment}
\label{Sec:exp}
\subsection{Experimental Setup}
\noindent\textbf{Datasets.}
We conduct a comprehensive evaluation of our method across a variety of benchmarks. The main paper reports results on the Semantic Shift Benchmark (SSB)~\cite{vaze2022semantic}, which covers fine-grained datasets-CUB \cite{wah2011caltech}, Stanford Cars~\cite{krause20133d}, and FGVC-Aircraft~\cite{maji2013fine}-plus Oxford-Pet~\cite{parkhi2012cats} and the more challenging Herbarium19~\cite{tan2019herbarium}.
To gauge generalization on standard recognition tasks, we also include results on the generic benchmarks CIFAR-10/100\cite{krizhevsky2009learning} and ImageNet-100~\cite{deng2009imagenet} in the Appendix.
For all datasets, we follow the class split protocol of \cite{vaze2022generalized}, where a subset of classes is selected as the known (`Old') label set $\mathcal{Y}_l$. From these known classes, $50\%$ of the samples are used to construct the labelled set $\mathcal{D}_l$, and the remaining images with instances from novel classes form the unlabelled set $\mathcal{D}_u$. Dataset statistics are summarized in Tab.~\ref{tab:datasplit}.

\begin{wraptable}{R}{0.45\textwidth}
\centering
\caption{Overview of dataset, including the classes in the labelled and unlabelled sets ($|\mathcal{Y}_l|$, $|\mathcal{Y}_u|$) and counts of images ($|\mathcal{D}_l|$, $|\mathcal{D}_u|$). }
\setlength{\tabcolsep}{1mm}{
\resizebox{0.45\textwidth}{!}{
\begin{tabular}{lccccc}
    \toprule
    Dataset &Balance &$|\mathcal{D}_l|$&$|\mathcal{Y}_l|$&$|\mathcal{D}_u|$ &$|\mathcal{Y}_u|$\\
    \midrule
    CUB~\cite{wah2011caltech} &\ding{51} & 1.5K & 100 & 4.5K & 200 \\
    Stanford Cars~\cite{krause20133d} &\ding{51} & 2.0K & 98 & 6.1K & 196 \\
    FGVC-Aircraft~\cite{maji2013fine} & \ding{51} &1.7K & 50 & 5.0K & 100 \\
    Oxford-Pet~\cite{parkhi2012cats}    & \ding{51}   & $0.9$K      & $19$       & $2.7$K     & $37$ \\
    Herbarium19~\cite{tan2019herbarium} &\ding{55} & 8.9K & 341 & 25.4K & 683 \\
    \bottomrule
\end{tabular}
}
}
\label{tab:datasplit}
\end{wraptable}

\noindent\textbf{Evaluation metrics.}
We evaluate GCD performance using clustering accuracy (\textit{ACC}), following standard practice~\cite{vaze2022generalized}. Specifically, given ground-truth labels $\bm{y}_i$ and predicted labels $\hat{\bm{y}}_i$ for the unlabelled set $\mathcal{D}_u$, the \textit{ACC} is computed as:
\begin{equation}
    \textit{ACC}=\frac{1}{|\mathcal{D}_u|}\sum\limits_{i=1}^{|\mathcal{D}_u|}\mathbbm{1}(\bm{y}_i=\mathbf{h}(\hat{\bm{y}}_i)), 
\end{equation}
where $\mathbf{h}$ denotes the optimal one-to-one mapping between predicted clusters and true class labels. For a comprehensive evaluation, we report \textit{ACC} separately for all classes (`All'), known classes (`Old'), and novel classes (`New').

\noindent\textbf{Implementation details.}
Following prior works~\cite{RastegarECCV2024, wen2023parametric, vaze2022generalized}, we adopt the ViT-B backbone~\cite{dosovitskiy2020image}, initialized with pretrained weights from either DINO~\cite{caron2021emerging} or DINOv2~\cite{oquab2023dinov2}.
The model is trained for 200 epochs using a batch size of 128 and a cosine learning rate schedule, starting from an initial learning rate of $10^{-1}$ and decaying to $10^{-4}$. All experiments are performed on a single NVIDIA L40S GPU with 24GB of memory. More details are provided in Appendix.

\subsection{Main Results}

\begin{table*}[t]
\centering
\caption{Comparison of state-of-the-art GCD methods on SSB~\cite{vaze2022semantic} benchmark. Results are reported in \textit{ACC} across the `All', `Old' and `New' categories.
The highest and second-highest scores are indicated in \textbf{bold} and \underline{underline} respectively.}
\setlength{\tabcolsep}{2mm}
\resizebox{0.85\textwidth}{!}{
\begin{tabular}{lll
  >{\columncolor{my_blue}}ccc
  >{\columncolor{my_blue}}ccc
  >{\columncolor{my_blue}}ccc
  >{\columncolor{my_blue}}c}
\toprule
&&&\multicolumn{3}{c}{CUB}&\multicolumn{3}{c}{Stanford Cars}&\multicolumn{3}{c}{FGVC-Aircraft}&\multicolumn{1}{c}{Average}\\
\cmidrule(lr){4-6} \cmidrule(lr){7-9} \cmidrule(lr){10-12}
&Method&Venue&All&Old&New&All&Old&New&All&Old&New&All\\
\midrule
\multirow{23}{*}{\rotatebox{90}{\emph{DINOv1}}}
&$k$-means~\cite{macqueen1967some}&-&34.3&38.9&32.1 &12.8&10.6&13.8 &16.0&14.4&16.8 &21.1\\
&RankStats+~\cite{han2021autonovel}&\textit{ICLR20}&33.3&51.6&24.2 &28.3&61.8&12.1 &26.9&36.4&22.2 &29.5\\
&UNO+~\cite{fini2021unified}&\textit{ICCV21}&35.1&49.0&28.1 &35.5&70.5&18.6 &40.3&56.4&32.2 &37.0\\
&ORCA~\cite{cao2022open}&\textit{CVPR22}&35.3&45.6&30.2 &23.5&50.1&10.7 &22.0&31.8&17.1 &26.9\\
&GCD~\cite{vaze2022generalized}&\textit{CVPR22}&51.3&56.6&48.7 &39.0&57.6&29.9 &45.0&41.1&46.9 &45.1\\
&XCon~\cite{fei2022xcon}&\textit{BMVC22}&52.1&54.3&51.0 &40.5&58.8&31.7 &47.7&44.4&49.4 &46.8\\
&OpenCon~\cite{sun2022opencon}&\textit{TMLR23}&54.7&63.8&54.7 &49.1&78.6&32.7 &-&-&- &-\\
&PromptCAL~\cite{zhang2023promptcal}&\textit{CVPR23}&62.9&64.4&62.1 &50.2&70.1&40.6 &52.2&52.2&52.3 &55.1\\
&DCCL~\cite{pu2023dynamic}&\textit{CVPR23}&63.5&60.8&64.9 &43.1&55.7&36.2 &-&-&- &-\\
&GPC~\cite{Zhao_2023_ICCV}&\textit{ICCV23}&52.0&55.5&47.5 &38.2&58.9&27.4 &43.3&40.7&44.8 &44.5\\
&PIM~\cite{chiaroni2023parametric}&\textit{ICCV23}&62.7&75.7&56.2 &43.1&66.9&31.6 &-&-&-&- \\
&SimGCD~\cite{wen2023parametric}&\textit{ICCV23}&60.3&65.6&57.7 &53.8&71.9&45.0 &54.2&59.1&51.8 &56.1\\
&$\mu$GCD~\cite{vaze2023no} &\textit{NeurIPS23}&65.7&68.0&64.6 &56.5&68.1&{50.9} &53.8&55.4&53.0 &58.7\\
&InfoSieve~\cite{rastegar2023learn}&\textit{NeurIPS23}&\textbf{69.4}&\textbf{77.9}&\textbf{65.2} &55.7&74.8&46.4 &56.3&{63.7}&52.5 &60.5\\
&TIDA~\cite{wang2023discover}&\textit{NeurIPS23}&{54.7}&{72.3}&{46.2} &-&-&- &54.6&{61.3}&52.1 &-\\
&CiPR~\cite{hao2023cipr}&\textit{TMLR24} &57.1&58.7&55.6 &47.0&61.5&40.1 &-&-&- &-\\
&SPTNet~\cite{wang2024sptnet}&\textit{ICLR24} &65.8&68.8&\underline{65.1} &\underline{59.0}&{79.2}&49.3 &{59.3}&61.8&{58.1} &\underline{61.4}\\
&Yang~\emph{et al.}~\cite{yang2024learning}&\textit{ECCV24} &61.3&60.8&62.1 &44.3&58.2&39.1 &-& -&- &-\\
&AMEND~\cite{Banerjee_2024_WACV}&\textit{WACV24} &64.9&\underline{75.6}&59.6 &52.8&61.8&48.3 &56.4& {73.3}&48.2 &\\
&LegoGCD~\cite{Cao_2024_CVPR}&\textit{CVPR24}&63.8&71.9&59.8 &57.3&{75.7}&48.4 &{55.0}&61.5&51.7&58.7\\ 
&MSGCD~\cite{DUAN2025103020}&\textit{IF25}&  {63.6}&70.7&{60.0} &57.7&75.5&49.9 &56.4&\underline{64.1}&52.6 &59.2\\
&DebGCD~\cite{liu2025debgcd} &\textit{ICLR25}&\underline{66.3}&{71.8}&63.5 &\textbf{65.3}&\textbf{81.6}&\underline{57.4} &\underline{61.7}&\underline{63.9}&\textbf{60.6} &{64.4}\\
\rowcolor{my_green}
&\textbf{Ours}&-&66.2&{72.1}&63.2&\textbf{65.3}&\underline{79.3}&\textbf{58.5}&\textbf{62.0}&\textbf{65.3}&\underline{60.4}&\textbf{64.5}\\
\midrule
\multirow{8}{*}{\rotatebox{90}{\emph{DINOv2}}}
&$k$-means~\cite{macqueen1967some}&-&67.6&60.6&71.1 &29.4&24.5&31.8 &18.9&16.9&19.9 &38.6\\
&GCD~\cite{vaze2022generalized}&\textit{CVPR22}&71.9&71.2&72.3 &65.7&67.8&64.7 &55.4&47.9&59.2& 64.3\\
&CiPR~\cite{hao2023cipr}&\textit{TMLR24}&\textbf{78.3}&73.4&\textbf{80.8} &66.7&77.0&61.8 &59.2&65.0&56.3 &68.1\\
&SimGCD~\cite{wen2023parametric}&\textit{ICCV23}&71.5&{78.1}&68.3 &71.5&81.9&66.6 &63.9&{69.9}&60.9 &69.0\\
&$\mu$GCD~\cite{vaze2023no}&\textit{NeurIPS23}&74.0&75.9&73.1 &\underline{76.1}&\textbf{91.0}&{68.9} &{66.3}&68.7&{65.1} &{72.1}\\
&SPTNet~\cite{wang2024sptnet}&\textit{ICLR24} &76.3&\underline{79.5}&74.6 &-&-&- &-&-&- &-\\
&{DebGCD~\cite{liu2025debgcd} }&\textit{ICLR25}&\underline{77.5}&\textbf{80.8}&{75.8} &{75.4}&{87.7}&\underline{69.5} &\underline{71.9}&\textbf{76.0}&\underline{69.8} &\underline{74.9}\\
\rowcolor{my_green}
&\textbf{Ours}&-&76.7&78.3&\underline{75.9}&\textbf{77.7}&\underline{88.7}&\textbf{72.4}&\textbf{74.6}&\underline{73.2}&\textbf{75.3}&\textbf{76.3}\\
\bottomrule
\end{tabular}
}
\label{tab:ssb}
\end{table*}

\begin{wraptable}{R}{0.4\textwidth}
\centering
\vspace{-16pt}
\caption{Comparison with state-of-the-art GCD methods on Herbarium19~\cite{tan2019herbarium} and Oxford-Pet~\cite{parkhi2012cats} on DINOv1.} 
\setlength{\tabcolsep}{1mm}{
\resizebox{0.4\textwidth}{!}{
\begin{tabular}{l>{\columncolor{my_blue}}ccc>{\columncolor{my_blue}}ccc}
    \toprule
     &\multicolumn{3}{c}{Oxford-Pet}&\multicolumn{3}{c}{Herbarium19}\\
    \cmidrule(lr{1em}){2-4} \cmidrule(lr{1em}){5-7} 
 Method&All&Old&New&All&Old&New\\ 
    \hline
    $k$-means~\cite{macqueen1967some}&77.1&70.1&80.7 &13.0&12.2&13.4 \\
    RankStats+~\cite{han2021autonovel}&-&-&- &27.9&55.8&12.8 \\
    UNO+~\cite{fini2021unified}&-&-&- &28.3&53.7&14.7 \\
    ORCA~\cite{cao2022open}&-&-&- &24.6&26.5&23.7 \\
    GCD~\cite{vaze2022generalized}&80.2&85.1&77.6 &35.4&51.0&27.0\\
    XCon~\cite{fei2022xcon}&86.7&\underline{91.5}&84.1 &-&-&-\\
    OpenCon~\cite{sun2022opencon}&-&-&- &39.3&58.9&28.6\\
    DCCL~\cite{pu2023dynamic}&88.1&88.2&88.0 &-&-&- \\
    SimGCD~\cite{wen2023parametric}&{91.7}&83.6&\underline{96.0} &44.0&58.0&36.4 \\
    $\mu$GCD~\cite{vaze2023no}&-&-&- &\underline{45.8}&\textbf{61.9}&\underline{37.2} \\
    InfoSieve~\cite{rastegar2023learn}&90.7&\textbf{95.2}&88.4 &40.3&59.0&30.2 \\ 
    {DebGCD~\cite{liu2025debgcd} }&\textbf{93.0}&86.4&\textbf{96.5} &{44.7}&\underline{59.}4&{36.8}\\
    \hline
    \textbf{Ours}&\underline{92.9}&88.9&95.0&\bf{46.9}&45.8&\bf{48.2}\\
    \bottomrule
\end{tabular}
}
\vspace{-22pt}
}
\label{tab:add}
\end{wraptable}

We present benchmark results of our method and compare it with nineteen state-of-the-art techniques in GCD  as well as three robust baselines derived from novel category discovery.
All methods are based on the DINO~\cite{caron2021emerging} and DINOv2~\cite{oquab2023dinov2} pre-trained backbone. 
This comparative evaluation encompasses performance on the fine-grained SSB benchmark~\cite{vaze2022semantic}, Oxford-Pet~\cite{parkhi2012cats} and Herbarium19~\cite{tan2019herbarium}, as shown in Tab.~\ref{tab:ssb} and Tab.~\ref{tab:add}.


As shown in Tab.~\ref{tab:ssb}, our method consistently achieves state-of-the-art performance on the SSB benchmark~\cite{vaze2022semantic} based on both DINO~\cite{caron2021emerging} and DINOv2~\cite{oquab2023dinov2} pretrained backbones.
Specifically, under the DINOv2 setting, our approach reaches an average `All' accuracy of $76.3\%$, outperforming the previous best method, DebGCD~\cite{liu2025debgcd}, by $1.4\%$ margin. 
Our framework demonstrates strong and stable improvements on both the Stanford Cars~\cite{krause20133d} and FGVC-Aircraft~\cite{maji2013fine} datasets, achieving the highest accuracy under both backbone settings. 
This highlights the effectiveness of our semantic-guided hierarchical design and contrastive learning strategy, particularly in domains where the semantic hierarchy aligns closely with the underlying structure of man-made categories, such as vehicles and aircraft.
On the CUB~\cite{wah2011caltech} dataset, although our method slightly lags behind DebGCD~\cite{liu2025debgcd} and the non-parametric method InfoSieve~\cite{rastegar2023learn},  we attribute this gap to the nature of bird taxonomy based on human-annotated semantics, which may introduce inconsistencies absent in more systematically defined hierarchies like those in artificial object domains.

As shown in Tab.~\ref{tab:add}, our method achieves competitive performance on the relatively easier Oxford-Pet dataset~\cite{parkhi2012cats}, outperforming the baseline. More notably, on the more challenging Herbarium19 benchmark~\cite{tan2019herbarium}, it sets a new state-of-the-art by surpassing the previous best method, $\mu$GCD~\cite{vaze2023no}, by $1.1\%$ on the `All' accuracy. These results highlight the robustness of our approach across both simple and complex open-world discovery scenarios.

\subsection{Analysis}

\noindent\textbf{Component Analysis.}
We conduct ablation studies to analyse the contributions of each major component in our framework: Hierarchical Learning, Consistency Self-Distillation, and Hierarchical Semantic-Guided Soft Contrastive Learning (HSCL). As shown in Tab.~\ref{tab:ablation}, we report results on the SSB benchmark~\cite{vaze2022semantic}, including Stanford Cars~\cite{krause20133d}, CUB~\cite{wah2011caltech}, and FGVC-Aircraft~\cite{maji2013fine} datasets, evaluated over `All', `Old', and `New' categories.
Starting from the baseline trained solely with the GCD loss, we incrementally integrate the proposed components. Incorporating hierarchical learning alone (Row (1)) yields a modest improvement, particularly on the old categories. Adding consistency-based self-distillation (Row (2)) further improves alignment and stability, while semantic-guided HSCL (Row (3)) significantly boosts performance on novel classes by leveraging cross-instance semantic similarity. When all components are combined, the full framework achieves substantial gains with $11.5\%$ on Stanford Cars, $7.8\%$ on FGVC-Aircraft, and $5.7\%$ on CUB.
\setlength{\textfloatsep}{5pt}
\setlength{\floatsep}{5pt}
\setlength{\intextsep}{5pt}

\begin{table}[t] 
\centering
\caption{Ablations. The results regarding the different components in our framework on SSB Benchmark~\cite{vaze2022semantic}. \textit{ACC} of `All', `Old' and `New' categories are listed. Red numbers indicate the improvement over the baseline.}
\setlength{\tabcolsep}{1mm}
\renewcommand{\arraystretch}{0.95}
\resizebox{\linewidth}{!}{
\begin{tabular}{ccccccccccccc}
    \toprule
    &\multirow{2}*{\shortstack{Hierarchical \\Learning}}&
    \multirow{2}*{\shortstack{Consistency \\ Self Distillation}}&
    \multirow{2}*{\shortstack{Semantic-guided\\ HSCL}}&
    \multicolumn{3}{c}{SCars}&
    \multicolumn{3}{c}{CUB}&
    \multicolumn{3}{c}{Aircraft}\\
    \cmidrule(lr{1em}){5-7} \cmidrule(lr{1em}){8-10}\cmidrule(lr{1em}){11-13}
    &&&&All&Old&New &All&Old&New &All&Old&New \\
    \midrule
    baseline&\ding{55}&\ding{55}&\ding{55} &53.8&71.9&45.0&60.3&65.6 &57.7&54.2&59.1&51.8\\
    (1)&\ding{51}&\ding{55}&\ding{55} &57.5&67.1 &52.9&57.0&57.8&56.6&52.8&56.4&51.0\\
    (2)&\ding{51}&\ding{51}&\ding{55} &62.6&78.3 &55.0&57.8&56.6&57.5&57.0&63.5&53.8\\
    (3)&\ding{51}&\ding{55}&\ding{51} &64.4&77.5 &58.1&62.5&67.5&60.0&57.4&58.5&56.8\\
    \rowcolor{my_blue} Ours&\ding{51}&\ding{51}&\ding{51} &
    \bf{65.3}\small(\textcolor{red}{\textbf{+11.5}})&
    \bf{79.3}\small(\textcolor{red}{\textbf{+7.4}})&
    \bf{58.5}\small(\textcolor{red}{\textbf{+13.5}})&
    \bf{66.2}\small(\textcolor{red}{\textbf{+5.9}})&
    \bf{72.1}\small(\textcolor{red}{\textbf{+6.5}})&
    \bf{63.2}\small(\textcolor{red}{\textbf{+5.5}})&
    \bf{62.0}\small(\textcolor{red}{\textbf{+7.8}})&
    \bf{65.3}\small(\textcolor{red}{\textbf{+6.2}})&
    \bf{60.4}\small(\textcolor{red}{\textbf{+8.6}})\\
    \bottomrule
\end{tabular}
}
\label{tab:ablation}
\end{table}

\begin{wraptable}{R}{0.5\textwidth}
\centering
\vspace{-5pt}
\caption{Experimental results regarding consistency temperature $\tau_{c}$ and ratio $\lambda_s$ to control the soft negative ratio on the unlabelled set and validation set of Stanford Cars~\cite{krause20133d} dataset.}
\setlength{\tabcolsep}{3mm}{
\resizebox{0.5\textwidth}{!}{
\begin{tabular}{ccccccc}
        \toprule
        &\multicolumn{3}{c}{Unlabelled Set}&\multicolumn{3}{c}{Validation Set}\\
        \cmidrule(lr{1em}){2-4} \cmidrule(lr{1em}){5-7}
        Param.&All&Old&New&All&Old&New\\
        \midrule
        $\tau_c=0.5$&62.9&77.5&55.9 &65.3&77.4&53.6\\
         \rowcolor{my_blue}$\tau_c=0.75$&\bf{65.3}&79.3&\bf{58.5} &\bf{66.4}&77.3&\bf{55.9}\\
        $\tau_c=1.0$&61.6&73.9&55.7 &63.7&75.3&52.6\\
        $\tau_c=1.25$&62.8&\bf{79.5}&54.7 &65.2&\bf{78.4}&52.6\\
        \hline
        $\lambda_s=0.2$&63.6&78.9& 56.3&65.6&78.1&53.5\\
        $\lambda_s=0.4$&63.9&78.5&56.9 &65.2&78.0&52.9\\
        $\lambda_s=0.6$&64.7&\bf{80.8}&56.9 &66.3&{78.3}&54.6\\
        $\lambda_s=0.8$&64.4&78.1&57.8 &66.1&\bf{78.4}&54.2\\
         \rowcolor{my_blue}$\lambda_s=1.0$&\bf{65.3}&79.3&\bf{58.5} &\bf{66.4}&77.3&\bf{55.9}\\
        \bottomrule
        \end{tabular}
}
}
\label{tab:hyp}
\end{wraptable}

\noindent\textbf{Hyperparameter Tuning.}
In line with the practices in~\cite{wen2023parametric, vaze2022generalized}, we perform hyperparameter tuning using a held-out validation split from the labelled data. Specifically, we tune the consistency temperature $\tau_c$ and the soft negative controller $\lambda_s$ based on their performance on the Stanford Cars~\cite{krause20133d} dataset. Detailed results across different hyperparameter values, evaluated on both the unlabelled training set and the validation split, are provided to assess their impact on model performance.
As shown in Tab.~\ref{tab:hyp}, we conduct a detailed grid search over the consistency temperature $\tau_c$ and the soft negative controller $\lambda_s$ on the Stanford Cars dataset.
Notably, the trends across both evaluation sets are highly consistent, with optimal performance achieved when $\tau_c = 0.75$ and $\lambda_s = 1.0$. These settings yield the best balance between old and new class performance, highlighting the importance of carefully tuning both the consistency strength and the soft negative ratio in our framework.

\begin{wraptable}{R}{0.4\textwidth}
\centering
\vspace{-10pt}
\caption{Results on Scars with alternative semantic hierarchies (vehicle brand vs. vehicle type) with DINOv2 pretrained backbone.}
\setlength{\tabcolsep}{3mm}{
\resizebox{0.4\textwidth}{!}{
\begin{tabular}{cccc}
        \toprule
        &\multicolumn{3}{c}{Scars}\\
        \cmidrule(lr{1em}){2-4} 
        Param.&All&Old&New\\
        \midrule
        SimGCD~\cite{wen2023parametric}&71.5&81.9&66.6 \\
        $\mu$GCD~\cite{vaze2023no}&76.1&\bf{91.0}&68.9 \\
        DebGCD~\cite{liu2025debgcd}&75.4&87.7&69.5 \\
        \rowcolor{my_blue}SEAL(Vehicle Brand)&\bf{77.1}&89.0&\bf{71.3} \\
        \rowcolor{my_blue}SEAL(Vehicle Type)&\bf{77.7}&88.7&\bf{72.4} \\
        \bottomrule
        \end{tabular}
}
}
\label{tab:semanticDimension}
\end{wraptable}
\noindent\textbf{Semantic Dimensions.}
Semantic hierarchies are not restricted to a single dimension. To further demonstrate the flexibility of our framework, we additionally adopt LLM-generated labels along an alternative semantic dimension, \eg, complementing the vehicle type hierarchy (SUV/Van/Coupe) with a brand-based hierarchy (Audi/BMW).
Tab.~\ref{tab:semanticDimension} demonstrates that our approach achieves consistently strong performance under both semantic hierarchies.
This highlights the robustness and flexibility of our proposed use of semantic-guided hierarchies across different semantic dimensions, and underscores their importance for GCD, whether sourced from curated taxonomies, generated by LLMs, or defined along alternative semantic structures.

\begin{figure}[h]
\centering
\includegraphics[width = 1\textwidth]{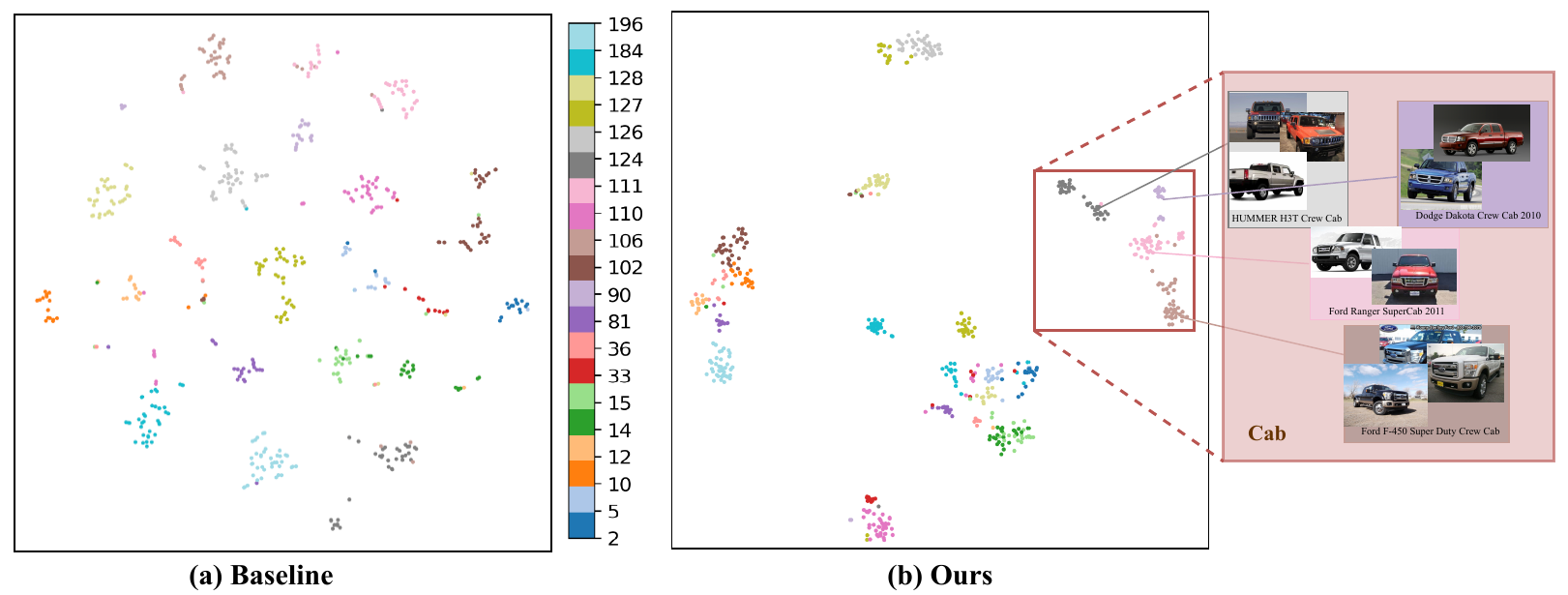}
\caption{
$t$-SNE visualization of 20 classes randomly sampled from the Stanford Cars~\cite{krause20133d} dataset.
}
\label{fig:tsne}
\end{figure}

\noindent\textbf{Visualization.}
We present a $t$-SNE~\cite{van2008visualizing} visualization comparing the feature representations learned by the baseline and ours. For clarity, we randomly select $20$ categories, including 10 from the `Old' set and $10$ from the `New' set. 
As shown in Fig.~\ref{fig:tsne}, our method yields tighter, better-separated clusters, indicating stronger inter-class discrimination. The zoomed view further reveals that the model preserves coarse-to-fine semantics: visually diverse subcategories within the broader \textit{`Cab'} group lie close together, yet each remains distinct. This confirms that our method captures hierarchical structure while retaining fine-grained separability.

\section{Conclusion}
In this paper, we introduce a semantic-aware hierarchical learning framework for Generalized Category Discovery, composed of three key components. 
\textit{Firstly}, we design a multi-task architecture that leverages naturally occurring semantic hierarchies to jointly learn coarse-to-fine category structures. 
\textit{Secondly}, we propose a Cross-Granularity Consistency (CGC) module that distils information between levels, eliminating label conflicts across the hierarchy.
\textit{Thirdly}, we develop a Hierarchical Soft Contrastive Learning strategy that incorporates semantic similarity into the contrastive objective, enabling fine-grained representation learning guided by structured semantic relationships. 
Our framework is theoretically motivated by information-theoretic principles, which highlight the benefit of incorporating hierarchical supervision to achieve tighter theoretical bounds. 
Evaluations on diverse fine-grained and generic benchmarks confirm consistent, state-of-the-art gains, demonstrating both theoretical soundness and strong empirical performance.

\section{Discussion}
\noindent\textbf{Limitations.}
It is important to acknowledge a limitation concerning the scale of validation within our study. The dataset used for model evaluation includes fewer than 700 instances, which constrains the breadth of category coverage.
This constrained sample size may not fully represent the diversity of categories encountered in real-world scenarios. Consequently, the application of our model to category discovery in more complex and varied situations could be restricted. 
Further research with larger, more comprehensive datasets is warranted to validate the robustness of our findings across a wider range of categories.

\noindent\textbf{Broader Impacts.}
This work presents a feasible method for discovering novel categories in unlabelled data, potentially benefiting a variety of applications such as robotics, healthcare, and autonomous driving, \textit{etc}.
However, there is a potential risk of misuse. The technology could be applied in surveillance to cluster unknown individuals, raising significant privacy concerns.
Therefore, it is imperative to carefully consider ethical guidelines and legal compliance to address concerns regarding individual privacy. Additionally, to mitigate potential negative social impacts, the development of robust security protocols and systems is crucial to protect sensitive information from cyberattacks and data breaches.


\bibliographystyle{plain}
\bibliography{neurips_2025}

\begin{thebibliography}{10}

\bibitem{achiam2023gpt}
Josh Achiam, Steven Adler, Sandhini Agarwal, Lama Ahmad, Ilge Akkaya, Florencia~Leoni Aleman, Diogo Almeida, Janko Altenschmidt, Sam Altman, Shyamal Anadkat, et~al.
\newblock Gpt-4 technical report.
\newblock {\em arXiv preprint arXiv:2303.08774}, 2023.

\bibitem{Banerjee_2024_WACV}
Anwesha Banerjee, Liyana~Sahir Kallooriyakath, and Soma Biswas.
\newblock Amend: Adaptive margin and expanded neighborhood for efficient generalized category discovery.
\newblock In {\em WACV}, 2024.

\bibitem{berthelot2019mixmatch}
David Berthelot, Nicholas Carlini, Ian Goodfellow, Nicolas Papernot, Avital Oliver, and Colin~A Raffel.
\newblock Mixmatch: A holistic approach to semi-supervised learning.
\newblock In {\em NeurIPS}, 2019.

\bibitem{boudiaf2020unifying}
Malik Boudiaf, J{\'e}r{\^o}me Rony, Imtiaz~Masud Ziko, Eric Granger, Marco Pedersoli, Pablo Piantanida, and Ismail~Ben Ayed.
\newblock A unifying mutual information view of metric learning: cross-entropy vs. pairwise losses.
\newblock In {\em ECCV}, 2020.

\bibitem{cao2022open}
Kaidi Cao, Maria Brbic, and Jure Leskovec.
\newblock Open-world semi-supervised learning.
\newblock In {\em ICLR}, 2022.

\bibitem{Cao_2024_CVPR}
Xinzi Cao, Xiawu Zheng, Guanhong Wang, Weijiang Yu, Yunhang Shen, Ke~Li, Yutong Lu, and Yonghong Tian.
\newblock Solving the catastrophic forgetting problem in generalized category discovery.
\newblock In {\em CVPR}, 2024.

\bibitem{caron2021emerging}
Mathilde Caron, Hugo Touvron, Ishan Misra, Herv{\'e} J{\'e}gou, Julien Mairal, Piotr Bojanowski, and Armand Joulin.
\newblock Emerging properties in self-supervised vision transformers.
\newblock In {\em ICCV}, 2021.

\bibitem{cendra2024promptccd}
Fernando~Julio Cendra, Bingchen Zhao, and Kai Han.
\newblock Promptccd: Learning gaussian mixture prompt pool for continual category discovery.
\newblock In {\em ECCV}, 2024.

\bibitem{Chang2021Labrador}
Dongliang Chang, Kaiyue Pang, Yixiao Zheng, Zhanyu Ma, Yi-Zhe Song, and Jun Guo.
\newblock Your “flamingo” is my “bird”: Fine-grained, or not.
\newblock In {\em CVPR}, 2021.

\bibitem{chapelle2009semi}
Olivier Chapelle, Bernhard Scholkopf, and Alexander Zien.
\newblock Semi-supervised learning (chapelle, o. et al., eds.; 2006)[book reviews].
\newblock {\em IEEE Transactions on Neural Networks}, 2009.

\bibitem{chen2021semi}
Xiaokang Chen, Yuhui Yuan, Gang Zeng, and Jingdong Wang.
\newblock Semi-supervised semantic segmentation with cross pseudo supervision.
\newblock In {\em CVPR}, 2021.

\bibitem{chiaroni2023parametric}
Florent Chiaroni, Jose Dolz, Ziko~Imtiaz Masud, Amar Mitiche, and Ismail Ben~Ayed.
\newblock Parametric information maximization for generalized category discovery.
\newblock In {\em ICCV}, 2023.

\bibitem{deng2009imagenet}
Jia Deng, Wei Dong, Richard Socher, Li-Jia Li, Kai Li, and Li~Fei-Fei.
\newblock Imagenet: A large-scale hierarchical image database.
\newblock In {\em CVPR}, 2009.

\bibitem{dosovitskiy2020image}
Alexey Dosovitskiy, Lucas Beyer, Alexander Kolesnikov, Dirk Weissenborn, Xiaohua Zhai, Thomas Unterthiner, Mostafa Dehghani, Matthias Minderer, Georg Heigold, Sylvain Gelly, et~al.
\newblock An image is worth 16x16 words: Transformers for image recognition at scale.
\newblock In {\em ICLR}, 2021.

\bibitem{du2021progressive}
Ruoyi Du, Jiyang Xie, Zhanyu Ma, Dongliang Chang, Yi-Zhe Song, and Jun Guo.
\newblock Progressive learning of category-consistent multi-granularity features for fine-grained visual classification.
\newblock {\em IEEE TPAMI}, 2021.

\bibitem{DUAN2025103020}
Yu~Duan, Zhanxuan Hu, Rong Wang, Zhensheng Sun, Feiping Nie, and Xuelong Li.
\newblock Mutual-support generalized category discovery.
\newblock {\em Information Fusion}, 2025.

\bibitem{fei2022xcon}
Yixin Fei, Zhongkai Zhao, Siwei Yang, and Bingchen Zhao.
\newblock Xcon: Learning with experts for fine-grained category discovery.
\newblock In {\em BMVC}, 2022.

\bibitem{fini2021unified}
Enrico Fini, Enver Sangineto, St{\'e}phane Lathuiliere, Zhun Zhong, Moin Nabi, and Elisa Ricci.
\newblock A unified objective for novel class discovery.
\newblock In {\em ICCV}, 2021.

\bibitem{gbif_download_2025}
{GBIF.org}.
\newblock Gbif occurrence download.
\newblock \url{https://doi.org/10.35035/d9pk-1162}, 2025.

\bibitem{gidaris2018dynamic}
Spyros Gidaris and Nikos Komodakis.
\newblock Dynamic few-shot visual learning without forgetting.
\newblock In {\em CVPR}, 2018.

\bibitem{girshick2015fast}
Ross Girshick.
\newblock Fast r-cnn.
\newblock In {\em ICCV}, 2015.

\bibitem{han2020automatically}
Kai Han, Sylvestre-Alvise Rebuffi, Sebastien Ehrhardt, Andrea Vedaldi, and Andrew Zisserman.
\newblock Automatically discovering and learning new visual categories with ranking statistics.
\newblock In {\em ICLR}, 2020.

\bibitem{han2021autonovel}
Kai Han, Sylvestre-Alvise Rebuffi, Sebastien Ehrhardt, Andrea Vedaldi, and Andrew Zisserman.
\newblock Autonovel: Automatically discovering and learning novel visual categories.
\newblock {\em IEEE TPAMI}, 2021.

\bibitem{han2019learning}
Kai Han, Andrea Vedaldi, and Andrew Zisserman.
\newblock Learning to discover novel visual categories via deep transfer clustering.
\newblock In {\em ICCV}, 2019.

\bibitem{han2024infomatch}
Qi~Han, Zhibo Tian, Chengwei Xia, and Kun Zhan.
\newblock Infomatch: Entropy neural estimation for semi-supervised image classification.
\newblock In {\em IJCAI}, 2024.

\bibitem{hao2023cipr}
Shaozhe Hao, Kai Han, and Kwan-Yee~K Wong.
\newblock Cipr: An efficient framework with cross-instance positive relations for generalized category discovery.
\newblock {\em TMLR}, 2024.

\bibitem{he2017mask}
Kaiming He, Georgia Gkioxari, Piotr Doll{\'a}r, and Ross Girshick.
\newblock Mask r-cnn.
\newblock In {\em ICCV}, 2017.

\bibitem{he2016deep}
Kaiming He, Xiangyu Zhang, Shaoqing Ren, and Jian Sun.
\newblock Deep residual learning for image recognition.
\newblock In {\em CVPR}, 2016.

\bibitem{he2025category}
Zhenqi He, Yuanpei Liu, and Kai Han.
\newblock Category discovery: An open-world perspective.
\newblock {\em arXiv preprint arXiv:2509.22542}, 2025.

\bibitem{transnuseg}
Zhenqi He, Mathias Unberath, Jing Ke, and Yiqing Shen.
\newblock Transnuseg: A lightweight multi-task transformer for nuclei segmentation.
\newblock In {\em MICCAI}, 2023.

\bibitem{jia2021joint}
Xuhui Jia, Kai Han, Yukun Zhu, and Bradley Green.
\newblock Joint representation learning and novel category discovery on single-and multi-modal data.
\newblock In {\em ICCV}, 2021.

\bibitem{joseph2022novel}
KJ~Joseph, Sujoy Paul, Gaurav Aggarwal, Soma Biswas, Piyush Rai, Kai Han, and Vineeth~N Balasubramanian.
\newblock Novel class discovery without forgetting.
\newblock In {\em ECCV}, 2022.

\bibitem{karbstein2024species}
Kevin Karbstein, Lara K{\o}sters, Ladislav Hoda{\v{c}}, Martin Hofmann, Elvira H{\"o}randl, Salvatore Tomasello, Natascha~D Wagner, Brent~C Emerson, Dirk~C Albach, Stefan Scheu, et~al.
\newblock Species delimitation 4.0: integrative taxonomy meets artificial intelligence.
\newblock {\em Trends in Ecology \& Evolution}, 2024.

\bibitem{krause20133d}
Jonathan Krause, Michael Stark, Jia Deng, and Li~Fei-Fei.
\newblock 3d object representations for fine-grained categorization.
\newblock In {\em ICCV workshop}, 2013.

\bibitem{krizhevsky2009learning}
Alex Krizhevsky, Geoffrey Hinton, et~al.
\newblock Learning multiple layers of features from tiny images.
\newblock 2009.

\bibitem{lang2024osr}
Nico Lang, Vésteinn Snæbjarnarson, Elijah Cole, Oisin Mac~Aodha, Christian Igel, and Serge Belongie.
\newblock From coarse to fine-grained open-set recognition.
\newblock In {\em CVPR}, 2024.

\bibitem{lin2014microsoft}
Tsung-Yi Lin, Michael Maire, Serge Belongie, James Hays, Pietro Perona, Deva Ramanan, Piotr Doll{\'a}r, and C~Lawrence Zitnick.
\newblock Microsoft coco: Common objects in context.
\newblock In {\em ECCV}, 2014.

\bibitem{liu2025debgcd}
Yuanpei Liu and Kai Han.
\newblock Debgcd: Debiased learning with distribution guidance for generalized category discovery.
\newblock In {\em ICLR}, 2025.

\bibitem{liu2025hyperbolic}
Yuanpei Liu, Zhenqi He, and Kai Han.
\newblock Hyperbolic category discovery.
\newblock In {\em CVPR}, 2025.

\bibitem{mace2004role}
Georgina~M Mace.
\newblock The role of taxonomy in species conservation.
\newblock {\em Philosophical Transactions of the Royal Society of London. Series B: Biological Sciences}, 2004.

\bibitem{macqueen1967some}
James MacQueen et~al.
\newblock Some methods for classification and analysis of multivariate observations.
\newblock In {\em Proceedings of the fifth Berkeley symposium on mathematical statistics and probability}, volume~1, pages 281--297. Oakland, CA, USA, 1967.

\bibitem{maji2013fine}
Subhransu Maji, Esa Rahtu, Juho Kannala, Matthew Blaschko, and Andrea Vedaldi.
\newblock Fine-grained visual classification of aircraft.
\newblock {\em arXiv preprint arXiv:1306.5151}, 2013.

\bibitem{miller1995wordnet}
George~A Miller.
\newblock Wordnet: a lexical database for english.
\newblock {\em Communications of the ACM}, 1995.

\bibitem{oord2018representation}
Aaron van~den Oord, Yazhe Li, and Oriol Vinyals.
\newblock Representation learning with contrastive predictive coding.
\newblock {\em ArXiv e-prints}, 2018.

\bibitem{oquab2023dinov2}
Maxime Oquab, Timoth{\'e}e Darcet, Th{\'e}o Moutakanni, Huy Vo, Marc Szafraniec, Vasil Khalidov, Pierre Fernandez, Daniel Haziza, Francisco Massa, Alaaeldin El-Nouby, et~al.
\newblock Dinov2: Learning robust visual features without supervision.
\newblock {\em arXiv preprint arXiv:2304.07193}, 2023.

\bibitem{parkhi2012cats}
Omkar~M Parkhi, Andrea Vedaldi, Andrew Zisserman, and CV~Jawahar.
\newblock Cats and dogs.
\newblock In {\em CVPR}, 2012.

\bibitem{peng2020mutual}
Jizong Peng, Marco Pedersoli, and Christian Desrosiers.
\newblock Mutual information deep regularization for semi-supervised segmentation.
\newblock In {\em Medical imaging with deep learning}, 2020.

\bibitem{pu2023dynamic}
Nan Pu, Zhun Zhong, and Nicu Sebe.
\newblock Dynamic conceptional contrastive learning for generalized category discovery.
\newblock In {\em CVPR}, 2023.

\bibitem{7583684}
Yanyun Qu, Li~Lin, Fumin Shen, Chang Lu, Yang Wu, Yuan Xie, and Dacheng Tao.
\newblock Joint hierarchical category structure learning and large-scale image classification.
\newblock {\em IEEE Transactions on Image Processing}, 2017.

\bibitem{rastegar2023learn}
Sarah Rastegar, Hazel Doughty, and Cees Snoek.
\newblock Learn to categorize or categorize to learn? self-coding for generalized category discovery.
\newblock In {\em NeurIPS}, 2023.

\bibitem{RastegarECCV2024}
Sarah Rastegar, Mohammadreza Salehi, Yuki~M Asano, Hazel Doughty, and Cees G~M Snoek.
\newblock Selex: Self-expertise in fine-grained generalized category discovery.
\newblock In {\em ECCV}, 2024.

\bibitem{ren2015faster}
Shaoqing Ren, Kaiming He, Ross Girshick, and Jian Sun.
\newblock Faster r-cnn: Towards real-time object detection with region proposal networks.
\newblock In {\em NeurIPS}, 2015.

\bibitem{simonyan2014very}
Karen Simonyan and Andrew Zisserman.
\newblock Very deep convolutional networks for large-scale image recognition.
\newblock In {\em ICLR}, 2015.

\bibitem{sun2022opencon}
Yiyou Sun and Yixuan Li.
\newblock Opencon: Open-world contrastive learning.
\newblock {\em TMLR}, 2022.

\bibitem{tan2019herbarium}
Kiat~Chuan Tan, Yulong Liu, Barbara Ambrose, Melissa Tulig, and Serge Belongie.
\newblock The herbarium challenge 2019 dataset.
\newblock {\em arXiv preprint arXiv:1906.05372}, 2019.

\bibitem{van2008visualizing}
Laurens Van~der Maaten and Geoffrey Hinton.
\newblock Visualizing data using t-sne.
\newblock {\em JMLR}, 2008.

\bibitem{vaze2022generalized}
Sagar Vaze, Kai Han, Andrea Vedaldi, and Andrew Zisserman.
\newblock Generalized category discovery.
\newblock In {\em CVPR}, 2022.

\bibitem{vaze2022semantic}
Sagar Vaze, Kai Han, Andrea Vedaldi, and Andrew Zisserman.
\newblock The semantic shift benchmark.
\newblock In {\em ICML workshop}, 2022.

\bibitem{vaze2023no}
Sagar Vaze, Andrea Vedaldi, and Andrew Zisserman.
\newblock No representation rules them all in category discovery.
\newblock In {\em NeurIPS}, 2023.

\bibitem{wah2011caltech}
Catherine Wah, Steve Branson, Peter Welinder, Pietro Perona, and Serge Belongie.
\newblock The caltech-ucsd birds-200-2011 dataset.
\newblock 2011.

\bibitem{wang2024sptnet}
Hongjun Wang, Sagar Vaze, and Kai Han.
\newblock Sptnet: An efficient alternative framework for generalized category discovery with spatial prompt tuning.
\newblock In {\em ICLR}, 2024.

\bibitem{wang2024hilo}
Hongjun Wang, Sagar Vaze, and Kai Han.
\newblock Hilo: A learning framework for generalized category discovery robust to domain shifts.
\newblock In {\em ICLR}, 2025.

\bibitem{wang2020solo}
Xinlong Wang, Tao Kong, Chunhua Shen, Yuning Jiang, and Lei Li.
\newblock Solo: Segmenting objects by locations.
\newblock In {\em ECCV}, 2020.

\bibitem{wang2023discover}
Yu~Wang, Zhun Zhong, Pengchong Qiao, Xuxin Cheng, Xiawu Zheng, Chang Liu, Nicu Sebe, Rongrong Ji, and Jie Chen.
\newblock Discover and align taxonomic context priors for open-world semi-supervised learning.
\newblock In {\em NeurIPS}, 2023.

\bibitem{9609630}
Xiu-Shen Wei, Yi-Zhe Song, Oisin~Mac Aodha, Jianxin Wu, Yuxin Peng, Jinhui Tang, Jian Yang, and Serge Belongie.
\newblock Fine-grained image analysis with deep learning: A survey.
\newblock {\em IEEE TPAMI}, 2022.

\bibitem{wen2023parametric}
Xin Wen, Bingchen Zhao, and Xiaojuan Qi.
\newblock Parametric classification for generalized category discovery: A baseline study.
\newblock In {\em ICCV}, 2023.

\bibitem{Wu_2024_CVPR}
Tz-Ying Wu, Chih-Hui Ho, and Nuno Vasconcelos.
\newblock Protect: Prompt tuning for taxonomic open set classification.
\newblock In {\em CVPR}, 2024.

\bibitem{xu2021end}
Mengde Xu, Zheng Zhang, Han Hu, Jianfeng Wang, Lijuan Wang, Fangyun Wei, Xiang Bai, and Zicheng Liu.
\newblock End-to-end semi-supervised object detection with soft teacher.
\newblock In {\em ICCV}, 2021.

\bibitem{yang2024learning}
Fengxiang Yang, Nan Pu, Wenjing Li, Zhiming Luo, Shaozi Li, Nicu Sebe, and Zhun Zhong.
\newblock Learning to distinguish samples for generalized category discovery.
\newblock In {\em ECCV}, 2024.

\bibitem{zhang2023promptcal}
Sheng Zhang, Salman Khan, Zhiqiang Shen, Muzammal Naseer, Guangyi Chen, and Fahad~Shahbaz Khan.
\newblock Promptcal: Contrastive affinity learning via auxiliary prompts for generalized novel category discovery.
\newblock In {\em CVPR}, 2023.

\bibitem{hierarchicalContrastiveLearning}
Shu Zhang, Ran Xu, Caiming Xiong, and Chetan Ramaiah.
\newblock Use all the labels: A hierarchical multi-label contrastive learning framework.
\newblock In {\em CVPR}, 2022.

\bibitem{zhao2021novel}
Bingchen Zhao and Kai Han.
\newblock Novel visual category discovery with dual ranking statistics and mutual knowledge distillation.
\newblock In {\em NeurIPS}, 2021.

\bibitem{Zhao_2023_ICCV}
Bingchen Zhao, Xin Wen, and Kai Han.
\newblock Learning semi-supervised gaussian mixture models for generalized category discovery.
\newblock In {\em ICCV}, 2023.

\bibitem{ncl}
Zhun Zhong, Enrico Fini, Subhankar Roy, Zhiming Luo, Elisa Ricci, and Nicu Sebe.
\newblock Neighborhood contrastive learning for novel class discovery.
\newblock In {\em CVPR}, 2021.

\end{thebibliography}



\clearpage
\appendix
\section*{Appendix}
{\hypersetup{linkcolor=black}
\startcontents[sections]
\printcontents[sections]{l}{1}{\setcounter{tocdepth}{2}}
}
\clearpage

\setcounter{table}{0} 
\setcounter{figure}{0} 
\setcounter{page}{1}
\renewcommand{\thetable}{A\arabic{table}}  
\section{ Theoretical Perspective}
In this section, we provide an information-theoretic proof motivating the design of \textbf{SEAL}.

\subsection{Notations \& Definitions}
\textit{Mutual Information} (MI) quantifies the reduction in uncertainty of one random variable given knowledge of another. We have the following definitions for MI:

\textbf{Definition.} The MI between two continuous random variables $X$ and $Y$ is formulated as.
\begin{equation}
    I(X;Y) =
\iint_{ X\times Y}
p_{XY}(x,y)\,
\log\!\frac{p_{XY}(x,y)}{p_X(x)\,p_Y(y)}
\,\mathrm d y\,\mathrm d x
\end{equation}

\textbf{Definition.} The MI between two discrete random variables $X$ and $Y$ is formulated as.
\begin{equation}
    I(X;Y) =
    \sum_{x,y}
    p_{XY}(x,y)\
    \log\!\frac{p_{XY}(x,y)}{p_X(x)\,p_Y(y)}
\end{equation}

 The notation we used and the related formulas are given in Tab.~\ref{table:notations}
\begin{table}[ht]
    \caption{Definition of the random variables and information measures used in this paper.}
    \label{table:notations}
    \centering
    \begin{tabular}{lc}
    \multicolumn{2}{c}{General} \\
    \toprule
    Labelled dataset & $\mathcal{D}_l=\{(\bm{x}_i, y_i)\}_{i=1}^n$ \\
    \midrule
    Unlabelled dataset & $\mathcal{D}_u=\{(\bm{x}_i, y_i)\}_{i=1}^m$ \\
    \midrule
    Image data space & $\mathcal{X}$ \\
    \midrule
    Embedded feature space & $Z \subset \mathbb{R}^d$ \\
    \midrule
    Label/Prediction space & $\mathcal{Y}/\mathcal{\hat{Y}} \subset \mathbb{R}^K$ \\
    \midrule
    Euclidean distance & $D_{ij} = \left\| \bm{x}_i - \bm{x}_j \right\|_2$ \\
    \midrule
    Cosine distance & $D_{cos_{ij}} = \frac{\bm{x}_i^\top  \bm{x}_j}{\left\| \bm{x}_i \right\| \left\| \bm{x}_j \right\|}$ \\
    \bottomrule
    \end{tabular}
    \qquad
    \begin{tabular}{lc}
    \multicolumn{2}{c}{Model} \\
    \toprule
    Encoder & $f_{\bm{\theta}}: \mathcal{X} \rightarrow{Z}$ \\
    \midrule
    Soft-classifier & $\mathcal{H} : \mathcal{Z} \rightarrow{[0, 1]^K}$ \\
    \bottomrule \\
    \multicolumn{2}{c}{Random variables (RVs)} \\
    \toprule
    Data &  ${X}=(X_l,X_u)$, ${Y}=(Y_l, Y_u)$ \\
    \midrule
    Embedding &  $Z|\mathcal{X} \sim f_{\bm{\theta}}(\mathcal{X})$ \\
    \midrule 
    Prediction &  $\hat{Y}|Z \sim \mathcal{H}_({Z})$ \\
    \bottomrule
    \end{tabular}
    \begin{tabular}{lc}
    \multicolumn{2}{c}{\rule{0pt}{4ex}Information measures} \\
    \toprule
    Mutual information between ${Z}$ and $Y$ & $\mathcal{I}({Z};Y) \coloneqq \mathcal{H}({Y}) - \mathcal{H}({Y}|{Z})$  \\
    \midrule
    Entropy of $Y$ & $ \mathcal{H}({Y})\coloneqq \mathbb{E}_{p_{Y}}\left[-\log p_Y (Y)\right]$ \\
    \midrule
    Conditional entropy of $Y$ given $Z$  & $ \mathcal{H}({Y|{Z}})\coloneqq \mathbb{E}_{p_{Y{Z}}} \left[-\log {p}_{Y|{Z}} (Y|{Z})\right]$ \\
    \midrule
    Cross entropy (CE) between $Y$ and $\widehat{Y}$  & $\mathcal{H}(Y; \widehat{Y}) \coloneqq \mathbb{E}_{p_Y} \left[-\log p_{\widehat{Y}} (Y)\right]$ \\
    \midrule
    Conditional CE given ${Z}$  & $\mathcal{H}(Y; \widehat{Y}|{Z}) \coloneqq \mathbb{E}_{p_{{Z}Y}} \left[-\log p_{\widehat{Y}|{Z}} (Y|{Z})\right]$ \\
    \bottomrule
    \end{tabular}
\end{table}


\subsection{Assumptions}
The following assumptions are made in our proof.

\textbf{A.1} Independent sampling between $(X_l, Y_l)$ and $(X_u, Y_u)$, which is written as $(X_l, Y_l) \perp\!\!\!\perp (X_u, Y_u) $.

\textbf{A.2} Same data distribution for $(X_l, Y_l)$ and $(X_u, Y_u)$ - the labelled and unlabelled data follow the same underlying data distribution \eg, same domain.

\textbf{A.3} The representation mapping $Z = f_\theta(X)$ is deterministic and per-sample independent given parameters $\theta$.

\subsection{Theoretical Motivations}
As shown in \cite{boudiaf2020unifying}, from the view of information theory, the optimization objective of discriminative tasks is equivalent to maximising the MI between the learned latent features ${Z}$ and $Y$, which is: 
\begin{equation}
    \max_\theta I_\theta (Z;Y) \Leftrightarrow \min_\theta -I_\theta (Z;Y).
\end{equation}

While this objective operates under the closed-world assumption, which assumes the availability of all annotations for the training data-in the GCD setting, both labelled and unlabelled data are present during training. Therefore, we further decompose the learning objective for GCD as follows. 
Given the \textit{chain rule for MI:}$I\!\bigl(X;\,Y_{1},\dots,Y_{n}\bigr)
   \;=\;
   \sum_{i=1}^{n} I\!\bigl(X;\,Y_{i}\mid Y_{1:i-1}\bigr)$, we can extend $I_\theta(Z;Y)$ as:
\begin{equation}
    \begin{aligned}
I_\theta(Z;Y)&=I_\theta (Z_l,Z_u; \mathcal Y_l, \mathcal Y_u)\\
&= I_\theta(Z_l, Z_u;\, \mathcal Y_l) 
              + I_\theta(Z_l, Z_u;\, \mathcal Y_u \mid \mathcal Y_l) \\
              &=I_\theta(Z_l;\mathcal Y_l) + I_\theta(Z_u|Z_l ;\mathcal Y_l) + I_\theta(Z_l, Z_u;\, \mathcal Y_u \mid \mathcal Y_l) \\
              & = I_\theta(Z_l;\mathcal Y_l) +  I_\theta(Z_u|Z_l ;\mathcal Y_l)  +I_\theta(Z_l; \mathcal Y_u \mid \mathcal Y_l) + I_\theta(Z_u|Z_l;\mathcal Y_u \mid \mathcal Y_l),\\
\end{aligned}
\end{equation}
where as $(Z_l, Y_l) \perp\!\!\!\perp (Z_u, Y_u) \;\;\Longleftrightarrow\;\; p(z_l,y_l,z_u,y_u)=p(z_l,y_l)\,p(z_u,y_u)$, we have:
\begin{equation}
    \begin{aligned}
        &I(Z_l;\mathcal Y_u \mid \mathcal Y_l)
 =\mathbb E_{y_l}\!
   \Bigl[
      \operatorname{KL}\!\bigl(
        p(z_l,y_u \mid y_l)
        \,\big\|\,
        p(z_l \mid y_l)\,p(y_u \mid y_l)
      \bigr)
   \Bigr]\\
   & = E_{y_l}\!
   \Bigl[
      \operatorname{KL}\!\bigl(
        \frac{p(z_l,y_l,y_u)}{p(y_l)}
        \,\big\|\,
        p(z_l \mid y_l)\,p(y_u \mid y_l)
      \bigr)
   \Bigr]\\
   & = E_{y_l}\!
   \Bigl[
      \operatorname{KL}\!\bigl(
        \frac{p(z_l,y_l)p(y_u)}{p(y_l)}
        \,\big\|\,
        p(z_l \mid y_l)\,p(y_u \mid y_l)
      \bigr)
   \Bigr]\\
   &= E_{y_l}\!
   \Bigl[
      \operatorname{KL}\!\bigl(
        p(z_l \mid y_l)\,p(y_u)
        \,\big\|\,
        p(z_l \mid y_l)\,p(y_u \mid y_l)
      \bigr)
   \Bigr] \textit{(By Bayes Rule)}\\
   &= E_{y_l}\!
   \Bigl[
      \operatorname{KL}\!\bigl(
        p(z_l \mid y_l)\,p(y_u)
        \,\big\|\,
        p(z_l \mid y_l)\,p(y_u)
      \bigr)
   \Bigr] \textit{ (By Independency)}\\
   &=0.
    \end{aligned}
\end{equation}
As the two arguments of the KL divergence are identical, we finally have $I(Z_l;\mathcal Y_u |\mathcal Y_l)=0$.

Similarly, for $I_\theta(Z_u|Z_l;\mathcal Y_l)$, we have:
\begin{equation}
    \begin{aligned}
        &I(Z_u\mid Z_l ;\mathcal Y_l) = I(\mathcal Y_l;Z_u\mid Z_l)
 =\mathbb E_{z_l}\!
   \Bigl[
      \operatorname{KL}\!\bigl(
        p(y_l,z_u \mid z_l)
        \,\big\|\,
        p(y_l \mid z_l)\,p(z_u \mid z_l)
      \bigr)
   \Bigr]\\
   & = E_{z_l}\!
   \Bigl[
      \operatorname{KL}\!\bigl(
        \frac{p(y_l,z_l,z_u)}{p(z_l)}
        \,\big\|\,
        p(y_l \mid z_l)\,p(z_u \mid z_l)
      \bigr)
   \Bigr]\\
   & = E_{z_l}\!
   \Bigl[
      \operatorname{KL}\!\bigl(
        \frac{p(y_l,z_l)p(z_u)}{p(z_l)}
        \,\big\|\,
        p(y_l \mid z_l)\,p(z_u \mid z_l)
      \bigr)
   \Bigr]\\
   &= E_{z_l}\!
   \Bigl[
      \operatorname{KL}\!\bigl(
        p(y_l \mid z_l)\,p(z_u)
        \,\big\|\,
        p(y_l \mid z_l)\,p(z_u \mid z_l)
      \bigr)
   \Bigr] \textit{(By Bayes Rule)}\\
   &= E_{z_l}\!
   \Bigl[
      \operatorname{KL}\!\bigl(
        p(y_l \mid z_l)\,p(z_u)
        \,\big\|\,
        p(y_l \mid z_l)\,p(z_u)
      \bigr)
   \Bigr] \textit{ (By Independency)}\\
   &=0.
    \end{aligned}
\end{equation}
By the independency assumption, we have $I_\theta (Z_u|Z_l; \mathcal Y_u | \mathcal Y_l) = I(Z_u;\mathcal  Y_u)$.

Therefore, the optimization objective is decomposed to:
\begin{equation}
    \min_\theta -I_\theta(Z_l; \mathcal Y_l) -I_\theta(Z_u;\mathcal Y_u)
\end{equation}
We further introduce a weight factor $\beta$~\cite{peng2020mutual, han2024infomatch} to balance between the supervised and unsupervised part to formulate the final objective as: 
\begin{equation}
    \min_\theta -I_\theta(Z_l;\mathcal Y_l) -\beta I_\theta(Z_u;\mathcal Y_u), 
\end{equation}
where for $I_\theta(Z_u;\mathcal Y_u)$, $\mathcal Y_u$ is unknown, we introduce a variational label distribution based on model prediction $q_\theta (\mathcal Y_u|X_u)  \triangleq p_\theta(\hat{Y}_u|X_u)$ where $\hat{Y}$ is the softmaxed model prediction. By the data-processing inequality that information passes through a transformation, mutual information with the source cannot increase, we have $I_\theta (Z_u; \mathcal Y_u) \geq I_\theta (X_u; \hat{Y}_u)$. From which, we can rewrite:
\begin{equation}
    \min - I_\theta(Z_u;Y_u) \rightarrow \min_\theta -I_\theta (X_u; \hat{Y}_u) = \min_\theta -H_{\theta}\!\bigl(\hat{\mathcal{Y}_u}\bigr)
   +\;
   H_{\theta}\!\bigl(\hat{\mathcal{Y}_u}\mid \mathcal{X}_{u}\bigr).
\end{equation}

Thus, the overall optimization objective can be reformulated as: 
\begin{equation}
     \min_{\theta}\;
-I_{\theta}\!\bigl(Z_{l};\,\mathcal{Y}_{l}\bigr)
\;+\;
\beta\,
\Bigl[
   -H_{\theta}\!\bigl(\hat{\mathcal{Y}_u}\bigr)
   +\;
   H_{\theta}\!\bigl(\hat{\mathcal{Y}_u}\mid \mathcal{X}_{u}\bigr)
\Bigr],
\end{equation}
where $\beta$ is the weight factor to balance labelled and unlabelled parts.



Assuming the coarse-grained semantic hierarchical labels $\mathcal{Y}_l^{(1)},...,\mathcal{Y}_l^{(H-1)}$ are accessible, the objective naturally extends to:
\begin{equation}
    \min_{\theta}
   \Big\{\underbrace{-\,I_\theta\!\bigl(Z_l;\,\mathcal{Y}_l^{(1)},\dots ,\mathcal{Y}_l^{(H)}\bigr)}_{\text{supervised part}}
   +\underbrace{\beta \Bigl[ H_\theta\!\bigl(\hat{ Y}^{(1:H)}_u\mid \mathcal{X}_u\bigr)
   -\,H_\theta\!\bigl(\hat{ Y}^{(1:H)}_u\bigr)}_{\text{unsupervised part}}
\Bigr]\Big\}.
\end{equation}

By applying the \textit{chain rule}, we first decompose the supervised part as:
\begin{equation}
    \begin{aligned}
       & I_\theta\!\bigl(Z_l;\mathcal Y_l^{(1:H)}\bigr) = I_\theta\!\bigl(Z_l;\mathcal Y_l^{(H:1)}\bigr) = I_\theta\!\bigl(Z_l;\mathcal Y_l^{(H)}\bigr)
     + \sum_{h=1}^{H-1}
       I_\theta\!\bigl(
           Z_l;\mathcal Y_l^{(h)}
           \,\bigm|\,
           \mathcal Y_l^{(h+1)},\dots,\mathcal Y_l^{(H)}
       \bigr) \\
       & \geq I_\theta\!\bigl(Z_l;\mathcal Y_l^{(H)}\bigr) \text{ as }\sum_{h=1}^{H-1}
       I_\theta\!\bigl(
           Z_l;\mathcal Y_l^{(h)}
           \,\bigm|\,
           \mathcal Y_l^{(h+1)},\dots,\mathcal Y_l^{(H)}
       \bigr) \geq 0,
    \end{aligned}
\end{equation}
where $\sum_{h=1}^{H-1}
       I_\theta\!\bigl(
           Z_l;\mathcal Y_l^{(h)}
           \,\bigm|\,
           \mathcal Y_l^{(h+1)},\dots,\mathcal Y_l^{(H)}
       \bigr) \geq 0$ comes from the below. For $\forall$ $h \in \{1\cdots H-1\}$, {we have:
\begin{equation}
\begin{aligned}
&I_\theta\!\bigl(
           Z_l;\mathcal Y_l^{(h)}
           \,\bigm|\,
           \mathcal Y_l^{(h+1)},\dots,\mathcal Y_l^{(H)}
       \bigr)
   \;=\;\\
   &\mathbb{E}_{c\sim p(\mathcal Y_l^{(h+1):(H)})}
      \!\Bigl[
        \operatorname{KL}\!\bigl(
            p_{Z_l,\mathcal Y_l^{(h)}\mid \mathcal Y_l^{(h+1):(H)}=c}
            \,\big\|\,
            p_{Z_l\mid \mathcal Y_l^{(h+1):(H)}=c}\,p_{\mathcal Y_l^{(h)}\mid \mathcal Y_l^{(h+1):(H)}=c}
        \bigr)
      \Bigr] \\
      & \geq 0. \textit{ (Non-negativity of KL divergence)}
\end{aligned}
\end{equation}

Similarly, for the unsupervised part, we can also obtain:
\begin{equation}
\begin{aligned}
H_\theta\!\bigl(\hat{ Y}^{(1:H)}_u\mid \mathcal{X}_u\bigr)
\;-\;
H_\theta\!\bigl(\hat{ Y}^{(1:H)}_u\bigr)
&= -\,I_\theta\!\bigl(\mathcal{X}_u;\hat Y^{(H)}_u\bigr)
    \;-\;\sum_{h=1}^{H-1} I_\theta\!\bigl(\mathcal{X}_u;\hat Y^{(h)} \mid \hat Y^{(h+1:H)}_u\bigr) \\[2pt]
&\le\; -\,I_\theta\!\bigl(\mathcal{X}_u;\hat Y^{(H)}_u\bigr)
=-H_{\theta}\!\bigl(\hat{{Y}_u}\bigr)
   +\
   H_{\theta}\!\bigl(\hat{{Y}_u}\mid \mathcal{X}_{u}\bigr).
\end{aligned}
\end{equation}

From the above, we now have:
\begin{equation}
    \begin{aligned}
        &\min_{\theta}
   \Big\{-\,I_\theta\!\bigl(Z_l;\,\mathcal{Y}_l^{(1)},\dots ,\mathcal{Y}_l^{(H)}\bigr)
   +\beta \Bigl[ H_\theta\!\bigl(\hat{ Y}^{(1:H)}_u\mid \mathcal{X}_u\bigr)
   -\,H_\theta\!\bigl(\hat{ Y}^{(1:H)}_u\bigr)
\Bigr]\Big\} \leq \\
&\min_{\theta}\Big\{\;
-I_{\theta}\!\bigl(Z_{l};\,\mathcal{Y}_{l}\bigr)
\;+\;
\beta\,
\Bigl[
   -H_{\theta}\!\bigl(\hat{\mathcal{Y}_u}\bigr)
   +\;
   H_{\theta}\!\bigl(\hat{\mathcal{Y}_u}\mid \mathcal{X}_{u}\bigr)
\Bigr]\Big\},
    \end{aligned}
\end{equation}
where we can see that the semantic-guided hierarchies provide a tighter bound on the mutual information, which motivates us to introduce the semantic-guided hierarchical learning framework for GCD.

\newpage
\section{Additional Details}
\subsection{Additional Implementation Details}
\label{Sec:implementation}
We adopt the class splits of labelled (`Old') and unlabelled (`New') categories in~\cite{vaze2022generalized} for generic object recognition datasets (including CIFAR-10~\cite{krizhevsky2009learning} and CIFAR-100~\cite{krizhevsky2009learning}) and the fine-grained Semantic Shift Benchmark~\cite{vaze2022semantic} (comprising CUB~\cite{wah2011caltech}, Stanford Cars~\cite{krause20133d}, and FGVC-Aircraft~\cite{maji2013fine}), Oxford-Pet~\cite{parkhi2012cats} and Herbarium19~\cite{tan2019herbarium}. 
Specifically, for all these datasets except CIFAR-100, $50\%$ of all classes are selected as `Old' classes ($\mathcal{Y}_l$), while the remaining classes are treated as `New' classes ($\mathcal{Y}_u \backslash \mathcal{Y}_l$). For CIFAR-100, $80\%$ of the classes are designated as `Old' classes, while the remaining $20\%$ as `New' classes. 
Moreover, following~\cite{vaze2022generalized} and~\cite{wen2023parametric}, the model's hyperparameters are chosen based on its performance on a hold-out validation set, formed by the original test splits of labelled classes in each dataset. 
All experiments utilize the PyTorch framework on a workstation with Nvidia L40s GPUs. The models are trained with a batch size of 128 on a single GPU for all datasets.

For the hierarchical information required by our framework, we rely exclusively on publicly available taxonomies or well-established datasets rather than any manual annotation. 
For the fine-grained SSB benchmarks~\cite{vaze2022semantic}, we follow the closed-world hierarchies of \cite{Chang2021Labrador}: CUB~\cite{wah2011caltech} is organised into $13$ orders, $38$ families, and $200$ species; Stanford Cars~\cite{krause20133d} is structured into $9$ car types (\eg, \textit{`Cab'}, \textit{`SUV'}) and $196$ specific models; FGVC-Aircraft~\cite{maji2013fine} is arranged into $30$ makers (\eg, \textit{`Boeing'}, \textit{`Douglas'}), 70 families (\eg, \textit{`Boeing 767'}), and $100$ models. 
Oxford Pets~\cite{parkhi2012cats} is re-cast into a two-level hierarchy with the coarse level \textit{`Cat'} \textit{vs.} \textit{`Dog'}, while Herbarium19~\cite{tan2019herbarium} is grouped by coarser-grained genus using the GBIF botanical database~\cite{gbif_download_2025}. 
For generic benchmarks, CIFAR-10~\cite{krizhevsky2009learning} is split into the super-classes \textit{`Vehicle'} and \textit{`Animal'}, CIFAR-100~\cite{krizhevsky2009learning} adopts its built-in 20 super-classes, and ImageNet-100~\cite{deng2009imagenet} leverages the WordNet~\cite{miller1995wordnet} taxonomy to form coarse categories. 
All hierarchies are obtained via public code, openly accessible biological and lexical databases or can be generated by LLMs, ensuring that our experiments reflect realistic usage without bespoke curation.

\subsection{Additional Dataset Details}

\begin{table}[h]
\centering
\caption{Overview of datasets we use, including the classes in the labelled and unlabelled sets ($|\mathcal{Y}_l|$, $|\mathcal{Y}_u|$) and counts of images ($|\mathcal{D}_l|$, $|\mathcal{D}_u|$). The `FG' indicates whether the dataset is fine-grained.}
\setlength{\tabcolsep}{1mm}{
\resizebox{0.5\textwidth}{!}{
\begin{tabular}{lccccc}
    \toprule
    Dataset &FG &$|\mathcal{D}_l|$&$|\mathcal{Y}_l|$&$|\mathcal{D}_u|$ &$|\mathcal{Y}_u|$\\
    \midrule
    CIFAR-10~\cite{krizhevsky2009learning} &\ding{55} & 12.5K & 5 & 37.5K & 10 \\
    CIFAR-100~\cite{krizhevsky2009learning} &\ding{55} & 20.0K & 80 & 30.0K & 100 \\
    ImageNet-100~\cite{deng2009imagenet} &\ding{55} & 31.9K & 50 & 95.3K & 100 \\
    CUB~\cite{wah2011caltech} &\ding{51} & 1.5K & 100 & 4.5K & 200 \\
    Stanford Cars~\cite{krause20133d} &\ding{51} & 2.0K & 98 & 6.1K & 196 \\
    FGVC-Aircraft~\cite{maji2013fine} & \ding{51} &1.7K & 50 & 5.0K & 100 \\
    Oxford-Pet~\cite{parkhi2012cats}    & \ding{51}   & $0.9$K      & $19$       & $2.7$K     & $37$ \\
    Herbarium19~\cite{tan2019herbarium} &\ding{51} & 8.9K & 341 & 25.4K & 683 \\
    \bottomrule
\end{tabular}
}
}
\label{tab:datasplit_appendix}
\end{table}

We further introduce the details of the datasets used in our paper. The statistics for the commonly used datasets are summarized in Tab.~\ref{tab:datasplit_appendix}.

\noindent\textbf{Generic Datasets.}
    (1) \textit{ImageNet-$100$}~\cite{deng2009imagenet} is a widely used dataset for natural image classification in computer vision, which is constructed by randomly subsampling $100$ classes from ImageNet-$1$K. 
    (2) \textit{CIFAR-$10$ \& CIFAR-$100$}~\cite{krizhevsky2009learning} are both natural images sized in $32 \times 32$. CIFAR-$10$ contains $50,000$ images spanning across $10$ different classes and CIFAR-$100$ includes $100$ classes, with each class containing $500$ images. 

\noindent\textbf{Fine-grained Datasets.}
The most widely used fine-grained benchmark is SSB~\cite{vaze2022generalized}, which includes three datasets: CUB~\cite{wah2011caltech}, Stanford Cars (SCars)~\cite{krause20133d}, and FGVC Aircraft~\cite{maji2013fine}.
    (1) \textit{CUB}~\cite{wah2011caltech} is a widely used benchmark dataset for fine-grained visual classification tasks, particularly focused on bird species recognition. 
    (2) \textit{Stanford Cars}~\cite{krause20133d} is a large-scale dataset designed for fine-grained vehicle classification tasks. It contains $196$ different car models, primarily spanning various makes, models, and years.
    (3) \textit{FGVC-Aircraft}~\cite{maji2013fine} is a fine-grained visual classification dataset focused on aircraft recognition. It contains 10,000 images spanning 100 different aircraft model variants, with each image labelled by its corresponding model. 
    (4) \textit{Oxford-Pet}~\cite{parkhi2012cats} is a large, fine-grained dataset designed for pet image classification and segmentation tasks. 
    (5) \textit{Herbarium19}~\cite{tan2019herbarium} is a large-scale image collection focused on plant species identification, particularly for herbarium specimen recognition. 

\newpage
\section{Experiments under Realistic Situation}

Following the majority of the literature, we conduct experiments mainly using the ground-truth category numbers. 
In this section, we test \textbf{SEAL} under more realistic conditions where neither coarse-granularity labels nor the number of classes are known. We adopt the same constraints used in earlier GCD works~\cite{vaze2022generalized,wen2023parametric}: only the \emph{known} fine-grained classes are revealed.
We first estimate the total number of targeted-granularity categories with an off-the-shelf method~\cite{vaze2022generalized}. Next, we automatically derive coarse-level names and the fine-to-coarse mapping using ChatGPT-4o~\cite{achiam2023gpt} with the following prompt: \texttt{``\{Targeted-grained Category Names\}'' I provide these \{Number of known category\} fine-grained class names, please generate the corresponding coarse-grained labels for me.}
After obtaining the coarse-granularity labels, we run the estimator~\cite{vaze2022generalized} to infer the number of coarse categories.
We test under such realistic condition for one fine-grained dataset (Stanford Cars) and one generic datasets (CIFAR100) and report the estimated class number about different granularities in Tab.~\ref{tab:estk}.
We compare \textbf{SEAL} with SimGCD~\cite{wen2023parametric}, $\mu$GCD~\cite{vaze2023no}, and GCD~\cite{vaze2022generalized} in Tab.~\ref{tab:estk1}. Even in this realistic scenario with an unknown number of categories and automatically generated coarse-granularity labels, our method outperforms existing approaches across both datasets. These results demonstrate that \textbf{SEAL} can be effectively deployed without any manual access to higher-level labels or class counts, while still achieving state-of-the-art accuracy.

\begin{table}[h]
\centering
\caption{Estimated class numbers in the unlabelled data using the method proposed in \cite{vaze2022generalized} for both target granularity and coarse granularity.}
\setlength{\tabcolsep}{1.5mm}{
\resizebox{0.8\textwidth}{!}{
\begin{tabular}{lccccc}
    \toprule
    &SCars (Target)&SCars (Coarse)&CIFAR-100 (Target)&CIFAR-100 (Coarse)\\
    \midrule
    Ground-truth $K$&200&9&100&20\\
    Estimated $K$&231&9&100&20\\
    \bottomrule
\end{tabular}
}
}
\label{tab:estk}
\end{table}

\begin{table}[h]
\centering
\caption{Results under the realistic scenario where neither coarse-granularity labels nor the number of classes are known. The estimated class numbers in Tab.~\ref{tab:estk} are adopted for all methods.
}

\setlength{\tabcolsep}{2mm}{
\resizebox{0.5\textwidth}{!}{
\begin{tabular}{l>{\columncolor{my_blue}}ccc>{\columncolor{my_blue}}ccc}
    \toprule
     &\multicolumn{3}{c}{Stanford Cars}&\multicolumn{3}{c}{CIFAR-100}\\
    \cmidrule(lr{1em}){2-4} \cmidrule(lr{1em}){5-7} 
 Method&All&Old&New&All&Old&New\\ \hline
    GCD~\cite{vaze2022generalized}&35.0&56.0&24.8  &73.0&76.2&66.5 \\
    SimGCD~\cite{wen2023parametric}&49.1&65.1&41.3  &80.1&81.2&77.8 \\
    $\mu$GCD~\cite{vaze2023no}&56.3&66.8&{51.1} &-&-&- \\
    \hline
    \textbf{Ours}&\bf{62.4}&\bf{78.9}&\bf{54.5} &\textbf{82.1}&\textbf{81.7}&\textbf{83.0}\\
    \bottomrule
\end{tabular}
}
}
\label{tab:estk1}
\end{table}

\section{Analysis on using randomly generated coarse-level labels}
To further substantiate our motivation that incorrect hierarchies may introduce misleading supervision, we conduct an experiment in which the true coarse-level labels are replaced with randomly generated hierarchies.
Specifically, we evaluate under two settings: $100\%$ random and $50\%$ random.
As shown in Tab.~\ref{tab:random}, performance drops sharply across CUB, SCars, and Aircraft under both variants of randomly assigned hierarchical labels.
This observation indicates that our gains arise from the semantic alignment of the hierarchy, not from the mere presence of a hierarchical structure.

\begin{table}[h]
\centering
\caption{Ablation on randomly generated coarse-level labels.}
\setlength{\tabcolsep}{3mm}
\resizebox{0.8\textwidth}{!}{
\begin{tabular}{cccccccccc}
        \toprule
        &\multicolumn{3}{c}{CUB}&\multicolumn{3}{c}{SCars}&\multicolumn{3}{c}{Aircraft}\\
        \cmidrule(lr{1em}){2-4} \cmidrule(lr{1em}){5-7} \cmidrule(lr{1em}){8-10}
        &All&Old&New&All&Old&New&All&Old&New\\
        \midrule
        $100\%$ Random&30.3&31.6&29.7 &29.6&31.4&28.7&33.2&31.3&34.2\\
        $50\%$ Random&51.2&50.3&51.7 &48.5&50.1&47.7&40.7&39.6&41.3\\
         \rowcolor{my_blue}\textbf{SEAL}&\bf{66.2}&\bf{72.1}&\bf{63.2} &{65.3}&79.3&58.5&\bf{62.0}&\bf{65.3}&\bf{60.4}\\
        \bottomrule
        \end{tabular}
}
\label{tab:random}
\end{table}

\newpage
\section{Results on Generic Datasets}
\begin{table*}[h]
\centering
\caption{Comparison of state-of-the-art GCD methods on generic datasets. It includes CIFAR-10~\cite{krizhevsky2009learning}, CIFAR-100~\cite{krizhevsky2009learning}, ImageNet-100~\cite{deng2009imagenet}, and the average \textit{ACC} on All categories.}
\setlength{\tabcolsep}{2.0mm}
\resizebox{0.8\textwidth}{!}{
\begin{tabular}{ll
  >{\columncolor{my_blue}}ccc
  >{\columncolor{my_blue}}ccc
  >{\columncolor{my_blue}}ccc
  >{\columncolor{my_blue}}c}
\toprule
&&\multicolumn{3}{c}{CIFAR-10}&\multicolumn{3}{c}{CIFAR-100}&\multicolumn{3}{c}{ImageNet-100}&\multicolumn{1}{c}{Average}\\
\cmidrule(lr){3-5} \cmidrule(lr){6-8} \cmidrule(lr){9-11}
&Method&All&Old&New&All&Old&New&All&Old&New&All\\
\midrule
\multirow{15}{*}{\rotatebox{90}{\emph{DINOv1}}}
&$k$-means~\cite{macqueen1967some} &83.6&85.7&82.5 &52.0&52.2&50.8 &72.7&75.5&71.3 &69.4\\
&RankStats+~\cite{han2021autonovel} &46.8&19.2&60.5 &58.2&77.6&19.3 &37.1&61.6&24.8 &47.4\\
&UNO+~\cite{fini2021unified} &68.6&\textbf{98.3}&53.8 &69.5&80.6&47.2 &70.3&\textbf{95.0}&57.9 &69.5\\
&ORCA~\cite{cao2022open} &69.0&77.4&52.0 &73.5&\textbf{92.6}&63.9 &81.8&86.2&79.6 &74.8\\
&GCD~\cite{vaze2022generalized} &91.5&\underline{97.9}&88.2 &73.0&76.2&66.5 &74.1&89.8&66.3 &81.1\\
&XCon~\cite{fei2022xcon} &96.0&97.3&95.4 &74.2&81.2&60.3 &77.6&93.5&69.7 &82.6\\
&OpenCon~\cite{sun2022opencon} &-&-&- &-&-&- &84.0&93.8&81.2 &-\\
&PromptCAL~\cite{zhang2023promptcal} &\textbf{97.9}&96.6&\underline{98.5} &81.2&84.2&75.3 &83.1&92.7&78.3 &87.4\\
&DCCL~\cite{pu2023dynamic} &96.3&96.5&96.9 &75.3&76.8&70.2 &80.5&90.5&76.2 &84.0\\
&GPC~\cite{Zhao_2023_ICCV} &90.6&97.6&87.0 &75.4&\underline{84.6}&60.1 &75.3&93.4&66.7 &80.4\\
&SimGCD~\cite{wen2023parametric} &97.1&95.1&98.1 &80.1&81.2&77.8 &83.0&93.1&77.9 &86.7\\
&InfoSieve~\cite{rastegar2023learn}&94.8&97.7&93.4 &78.3&82.2&70.5 &80.5&93.8&73.8 &84.5\\
&CiPR~\cite{hao2023cipr} &\underline{97.7}&97.5&97.7 &{81.5}&82.4&{79.7} &80.5&84.9&78.3 &86.6 \\
&SPTNet~\cite{wang2024sptnet} &97.3&95.0&\textbf{98.6} &81.3&84.3&75.6 &\underline{85.4}&93.2&\underline{81.4} &88.0\\
&{DebGCD~\cite{liu2025debgcd} }&97.2&94.8&98.4 &\textbf{83.0}&\underline{84.6}&\underline{79.9} &\textbf{85.9}&\underline{94.3}&\textbf{81.6} &88.7\\
\rowcolor{my_green}&{\textbf{Ours}}&97.2&94.7&98.4 &\underline{82.1}&{81.7}&\textbf{83.0} &84.6&90.9&81.3 &88.0\\
\midrule
\multirow{6}{*}{\rotatebox{90}{\emph{DINOv2}}}
&GCD~\cite{vaze2022generalized}&97.8&\textbf{99.0}&97.1 &79.6&84.5&69.9 &78.5&89.5&73.0 &85.3 \\
&CiPR~\cite{hao2023cipr}&\textbf{99.0}&\underline{98.7}&99.2 &\textbf{90.3}&89.0&\textbf{93.1} &88.2&87.6&{88.5} &92.5 \\
&SimGCD~\cite{wen2023parametric} &98.7&96.7&\textbf{99.7} &88.5&\underline{89.2}&87.2 &89.9&95.5&87.1 &92.4\\
&SPTNet~\cite{wang2024sptnet} &-&-&- &-&-&- &{90.1}&\underline{96.1}&87.1 &-\\
&{DebGCD~\cite{liu2025debgcd} } &\underline{98.9}&97.5&\underline{99.6} &\underline{90.1}&\textbf{90.9}&{88.6} &\textbf{93.2}&\textbf{97.0}&\textbf{91.2} &94.1\\
\rowcolor{my_green}&{\textbf{Ours} }&\underline{98.9}&98.1&{99.3} &{89.8}&{90.4}&\underline{89.5} &\underline{91.3}&93.3&\underline{90.3} &93.3\\
\bottomrule
\end{tabular}
}
\label{tab:generic}
\end{table*}

Tab.~\ref{tab:generic} shows that \textbf{SEAL} remains effective even when only shallow hierarchies are available. With using both DINO~\cite{caron2021emerging} abd DINOv2~\cite{oquab2023dinov2} pre-trained backbones, SEAL surpasses the strong SimGCD~\cite{wen2023parametric} baseline on all three datasets-CIFAR-10, CIFAR-100 and ImageNet-100.
For generic datasets, they provide only coarse and heterogeneous groupings (\eg, \textit{`Animal'} \textit{vs.} \textit{`Vehicle'} in CIFAR-10), so the hierarchy does not converge to a common parent class. By contrast, fine-grained datasets like CUB~\cite{wah2011caltech} share a clear taxonomic root (\eg, the \textit{class} \textit{`Aves'} for all bird species), allowing our method to exploit deeper and more coherent semantic structure. 
Even under this less favourable condition, \textbf{SEAL} still delivers competitive performance, confirming the robustness of our hierarchical design.

\section{Analysis on the Depth of Semantic Hierarchies}
Sec.~\ref{Sec:implementation} notes that our framework uses different hierarchical depths depending on dataset availability. 
To quantify the effect of depth, we conduct an ablation study on the two datasets that provide three explicit levels, including CUB~\cite{wah2011caltech} and FGVC-Aircraft~\cite{maji2013fine}. 
For each dataset we compare: (i) a single-level baseline that uses only the target granularity, (ii) a two-level version that adds one parent level, and (iii) the full three-level setting adopted in the main paper. 
Tab.~\ref{tab:ablation_depth} shows that \textbf{SEAL} remains effective irrespective of the number of available semantic levels. When compared to the single-granularity baseline across both datasets, the incorporation of just one additional coarse-granularity level yields improvements of approximately $2\%$ and $4\%$. These results demonstrate the robustness of our design, which can leverage richer hierarchies when they are present, while still providing significant benefits regardless of the number of hierarchies utilized.

\begin{table}[h]
\centering
\caption{Ablations analysis on the depth of semantic hierarchies. \textit{ACC} of `All', `Old' and `New' categories on Stanford Cars and FGVC-Aircraft are listed.}
\setlength{\tabcolsep}{1mm}{
\resizebox{0.9\textwidth}{!}{
\begin{tabular}{ccccccccc}
    \toprule
&\multirow{2}*{\shortstack{Depth \\of Hierarchies}}&\multirow{2}*{\shortstack{Coarser\\ Hierarchy}}&\multicolumn{3}{c}{CUB}&\multicolumn{3}{c}{FGVC-Aircraft}\\
\cmidrule(lr{1em}){4-6} \cmidrule(lr{1em}){7-9}
&&&All&Old&New &All&Old&New \\
    \midrule
    (i) Baseline&1&-&60.3&65.6 &57.7&54.2&59.1&51.8\\
    (ii)&2&Family&63.5&73.9 &58.3&58.4&63.6&55.8\\
    (ii)&2&Order / Maker&62.3&72.3 &57.3&58.6&60.7&58.3\\
    \rowcolor{my_blue} (iii) SEAL&3&\shortstack{Order/Maker + Family}&\bf{66.2}\small(\textcolor{red}{\textbf{+5.9}})&\bf{72.1} \small(\textcolor{red}{\textbf{+6.5}})&\bf{63.2} \small(\textcolor{red}{\textbf{+5.5}})&\bf{62.0} \small(\textcolor{red}{\textbf{+7.8}})&\bf{65.3} \small(\textcolor{red}{\textbf{+6.2}})&\bf{60.4} \small(\textcolor{red}{\textbf{+8.6}})\\
    \bottomrule
    
\end{tabular}
}
}
\label{tab:ablation_depth}
\end{table}

\newpage
\section{Analysis of Computational Costs}
Tab.~\ref{tab:compute} presents a comprehensive analysis of the computational cost associated with our method compared to the SimGCD baseline~\cite{wen2023parametric} for the training stage. Despite incorporating additional multi-level supervision, our approach introduces minimal computational overhead during training. 
Specifically, the number of parameters increases by less than $9\%$ across all datasets (from approximately $630$ MB to $688$ MB), and the GFLOPs remain virtually identical, showing only a marginal increase from 17.59 to 17.60. 
Importantly, the training efficiency is largely preserved, with time per epoch on the unlabelled dataset increasing by no more than 0.7 seconds in all cases. These results clearly demonstrate that our framework achieves its performance improvements without sacrificing computational efficiency in training stage.
At inference, however, we discard the coarse-granularity branches and keep only the classifier for the target granularity. 
The cost breakdown in Tab.~\ref{tab:compute_test} reveals that our model actually uses fewer parameters than the SimGCD baseline~\cite{wen2023parametric}.
Runtime on the unlabelled test sets is reduced for all three datasets - Stanford Cars~\cite{krause20133d}, CUB~\cite{wah2011caltech}, and FGVC-Aircraft~\cite{maji2013fine}.
This economy stems from our design: a single MLP projector separates features across levels without enlarging the overall feature dimension, so the target-level head is compact at test time.
Consequently, our method introduces almost no overhead during training and even lowers the computational footprint at inference, while still boosting accuracy.
\begin{table}[h]
\centering
\caption{
Computational cost analysis with baseline during training.}
\setlength{\tabcolsep}{2mm}{
\resizebox{0.9\textwidth}{!}{
\begin{tabular}{lccccccccc}
    \toprule
     &\multicolumn{3}{c}{\# Params (MB)$\downarrow$}&\multicolumn{3}{c}{GFLOPs$\downarrow$}&\multicolumn{3}{c}{Time per Epoch (s)$\downarrow$}\\
    \cmidrule(lr{1em}){2-4} \cmidrule(lr{1em}){5-7} \cmidrule(lr{1em}){8-10} 
 Method&SCars&CUB&Aircraft&SCars&CUB&Aircraft&SCars&CUB&Aircraft  \\ \hline
    SimGCD~\cite{wen2023parametric}&630.9&630.9&630.6 &17.59&17.59&17.59 &59.2&25.0&74.9  \\
    Ours&660.5&688.0&688.1 &17.60&17.60&17.60 &59.8&25.1& 75.6\\
    \bottomrule
\end{tabular}
}
}

\label{tab:compute}
\end{table}

\begin{table}[h]
\centering
\caption{
Computational cost analysis with baseline at inference time.}
\setlength{\tabcolsep}{2mm}{
\resizebox{0.9\textwidth}{!}{
\begin{tabular}{lccccccccc}
    \toprule
     &\multicolumn{3}{c}{\# Params (MB)$\downarrow$}&\multicolumn{3}{c}{GFLOPs$\downarrow$}&\multicolumn{3}{c}{Time per Epoch (s)$\downarrow$}\\
    \cmidrule(lr{1em}){2-4} \cmidrule(lr{1em}){5-7} \cmidrule(lr{1em}){8-10} 
 Method&SCars&CUB&Aircraft&SCars&CUB&Aircraft&SCars&CUB&Aircraft  \\ \hline
    SimGCD~\cite{wen2023parametric}&630.9&630.9&630.6 &17.59&17.59&17.59 &56.9&34.1&53.1  \\
    Ours&629.0&627.6&627.5 &17.59&17.59&17.59 &56.8&34.0& 52.9 \\
    \bottomrule
\end{tabular}
}
}

\label{tab:compute_test}
\end{table}

\section{Analysis on the Curriculum Learning Schedule}
As introduced in Sec.4.2.3, we employ a linear decay schedule for $\lambda_c$. Tab.~\ref{tab:decay} reports an ablation study on the decay strategy for the curriculum weighting coefficient $\lambda_c$. We compare fixed values ($\lambda_c=0,0.5,1$), and exponential decay schedule, and our proposed SEAL with linear decay. The results consistently show that decaying schedules outperform fixed baselines, validating the effectiveness of progressively shifting focus from coarse semantic alignment to finer positional discrimination. In particular, SEAL achieves the best or comparable performance across three fine-grained datasets (CUB, Stanford-Cars, and Aircraft), demonstrating that the linear decay schedule provides a more stable and effective curriculum learning design.
\begin{table}[h]
\centering
\caption{Ablation on curriculum decay strategies.}
\setlength{\tabcolsep}{3mm}{
\resizebox{0.8\textwidth}{!}{
\begin{tabular}{cccccccccc}
        \toprule
        &\multicolumn{3}{c}{CUB}&\multicolumn{3}{c}{SCars}&\multicolumn{3}{c}{Aircraft}\\
        \cmidrule(lr{1em}){2-4} \cmidrule(lr{1em}){5-7} \cmidrule(lr{1em}){8-10}
        &All&Old&New&All&Old&New&All&Old&New\\
        \midrule
        $\lambda_c=0$&64.6&72.2&60.8 &64.5&80.8&56.6&59.3&61.8&58.0\\
        $\lambda_c=0.5$&65.1&72.8&61.3 &64.1&77.1&57.8&58.9&62.5&57.1\\
        $\lambda_c=1.0$&64.9&72.4&60.9 &63.2&80.2&55.0&58.7&65.4&55.3\\
        Exp Decay&66.1&71.7&63.3 &\bf{66.3}&\bf{81.2}&\bf{60.3}&61.2&63.1&60.2\\
         \rowcolor{my_blue}\textbf{SEAL}&\bf{66.2}&\bf{72.1}&\bf{63.2} &{65.3}&79.3&58.5&\bf{62.0}&\bf{65.3}&\bf{60.4}\\
        \bottomrule
        \end{tabular}
}
}
\label{tab:decay}
\end{table}





\newpage

\end{document}